\documentclass[12pt]{iopart}
\usepackage{algorithmic}
\usepackage{algorithm}
\usepackage{array}
\usepackage[caption=false,font=normalsize,labelfont=sf,textfont=sf]{subfig}
\usepackage{textcomp}
\usepackage{stfloats}
\usepackage{url}
\usepackage{verbatim}
\usepackage{graphicx}
\usepackage{color, soul}
\usepackage{multirow}
\usepackage{makecell}
\usepackage{placeins}
\usepackage{hyperref}

\usepackage{amsfonts}
\usepackage{siunitx}
\usepackage{booktabs}

\hypersetup{
    colorlinks=true,
    linkcolor=black,
    filecolor=black, 
    citecolor=black,
    urlcolor=blue,
    breaklinks=true
    }
\usepackage{silence}
\begin{document}
\bibliographystyle{iopart-num-mod}

\title{xEEGNet: Towards Explainable AI in EEG Dementia Classification}
\author{Andrea Zanola$^{1,2,*}$, Louis Fabrice Tshimanga$^{1,2,3}$, Federico Del Pup$^{1,2,3}$, Marco Baiesi$^{4,5}$ and Manfredo Atzori$^{1,2,6}$}
\address{$^1$ Department of Neuroscience, University of Padua, Padua, 35128, Italy}
\address{$^2$ Padua Neuroscience Center, Padua, 35128, Italy}
\address{$^3$ Department of Information Engineering, University of Padua, Padua, 35128, Italy}
\address{$^4$ Department of Physics and Astronomy, University of Padua, Padua, 35128, Italy}
\address{$^5$ INFN, Section of Padua, Padua, 35128, Italy}
\address{$^6$ Information Systems Institute, University of Applied Sciences Western Switzerland (HES-SO Valais), 3960 Sierre}
\address{* Author to whom any correspondence should be addressed.}
\eads{\mailto{andrea.zanola@studenti.unipd.it}, \mailto{manfredo.atzori@unipd.it}}

\begin{abstract}
\textit{Objective.}
This work presents xEEGNet, a novel, compact, and explainable neural network for EEG data analysis.
It is fully interpretable and reduces overfitting through a major parameter reduction.

\textit{Approach.}
As an applicative use case to develop our model, we focused on the classification of common dementia conditions, Alzheimer's and frontotemporal dementia, versus controls.
xEEGNet, however, is broadly applicable to other neurological conditions involving spectral alterations.
We used ShallowNet, a simple and popular model in the EEGNet family, as a starting point.
Its structure was analyzed and gradually modified to move from a ``black box" model to a more transparent one, without compromising performance.
The learned kernels and weights were analyzed from a clinical standpoint to assess their medical significance.
Model variants, including ShallowNet and the final xEEGNet, were evaluated using a robust Nested-Leave-N-Subjects Out cross-validation for unbiased performance estimates.
Variability across data splits was explained using the embedded EEG representations, grouped by class and set, with pairwise separability to quantify group distinction.
Overfitting was measured through training-validation loss correlation and training speed.

\textit{Main results.}
xEEGNet uses only 168 parameters, 200 times fewer than ShallowNet, yet retains interpretability, resists overfitting, achieves comparable median performance (-1.5\%), and reduces performance variability across splits.
This variability is explained by the embedded EEG representations: higher accuracy correlates with greater separation between test-set controls and Alzheimer's cases, without significant influence from the training data.

\textit{Significance.}
The capability of xEEGNet to filter specific EEG bands, learns band specific topographies and use the right EEG spectral bands for disease classification demonstrates its interpretability.
While big deep learning models are typically prioritized for performance, this study shows that smaller architectures like xEEGNet can be equally effective in pathology classification, using EEG data.

\end{abstract}

\vspace{2pc}

\maketitle

% CHAPTERS
\section{Introduction}\label{introduction}
% GENERAL INTRODUCTION
In recent years, Deep Learning (DL) has emerged as a powerful new tool for the analysis of electroencephalography (EEG) data \cite{SSL_EEG}. 
Its superiority over standard statistical techniques lies in the high expressiveness of these architectures, which are capable of learning highly nonlinear functions.
These functions, commonly expressed as $f:\mathcal{D}\rightarrow \mathcal{L}$, represent arbitrarily complex mappings from the input space, the dataset $\mathcal{D}$, to the set of labels (classes) $\mathcal{L}$ in a classification problem.
% INTEPRETABILITY ISSUE IS THE MAIN POINT
Although the performances of such algorithms on various EEG-based tasks have been demonstrated to be impressive and promising in many areas \cite{Yannick}, their applicability remains limited in real-life scenarios \cite{Limitations}.
Among other reasons, including decision bias, quality of input and data partitioning experimental setup, there are serious limitations due to the lack of interpretability and transparency of these architectures.
Over the years, these limitations have been consistently emphasized in the literature, as highlighted in several studies 2018 \cite{prob2018}, 2021 \cite{prob2021} and 2024 \cite{prob2024}.

Deep neural networks are often described as ``black box" models, but eXplainable AI (XAI) actively seeks to make them more interpretable \cite{TowardBestPractice}.
XAI is particularly important in the medical domain, where model performance is not the only factor of relevance; as noted in \cite{THELANCETRESPIRATORYMEDICINE2018801}, accuracy alone is not sufficient to inspire trust.
Citing Jacovi \etal~\cite{Jacovi}, by designing an AI that users can and will trust to interact with, AI can be safely deployed in society.

% WE CAN ADDRESS IT USING THIS NETWORK
Consequently, the main objective of this work is to transform a ``black box'' model into a ``white box'' one by progressively modifying the layers in order to increase the interpretability of the network, while maintaining the same performance in the task under consideration.
In this study, we adopt the definition of interpretability provided by Miller \etal~\cite{MILLER20191}, who describes it as the degree to which a human can understand the reasons behind a model's decision. 
Thus, the issue of interpretability can be solved, by understanding the analytical form of the function $f$ that maps input to output. 
For very deep architectures with many layers, it becomes nearly impossible to provide both an analytical function and a physical interpretation. 
ShallowConvNet \cite{shallow}, often referred to as ShallowNet, is one of the most compact architectures related to the EEGNet family \cite{eegnet}, a group of lightweight convolutional networks designed for efficient EEG signal classification.
Despite its simplicity, it shows competitive performances in various tasks \cite{eegprepro} with respect to other larger models, making it a suitable choice for this type of analysis.

% BACKGROUND & SIMILAR WORKS
% A SIMILAR WORK RESPECT THE INTERPRETABILITY
A similar approach was taken by Borra \etal~\cite{BORRA202055}, 2020, who proposed a new model called Sinc-ShallowNet, which shares similarities with ShallowNet. 
They provided interesting insights into the features learned by the network in the first layer (band-pass filters) and offered valuable suggestions on techniques to improve the interpretability of the extracted features in the second layer (use of the depthwise convolutional layer \cite{Xception}, introduced in the EEG domain by Lawhern \etal \cite{eegnet}, 2018).
They also detailed the spatial distribution of the most relevant and class-specific band-pass filters learned by Sinc-ShallowNet for a given subject.
%WAYS IN WHICH OUR WORK DIFFERS
However, first, they restricted their description of the architecture to general $f$ notation; second, even if their suggestions and modifications are useful, they focused primarily on performance metrics derived from the test set leaving behind other aspects like interpretability; and third, they only used movement execution and imagination datasets, such as the famous, publicly available BCI-IV2a \cite{BCIdataset}.
%They did not perform any inter-subject analyses (which are instead fundamental for most tasks apart from BCI in EEG data analytics XXXRef LOSO), and did not include any dataset related to pathologic conditions.
Differently, this paper focuses on inter-subject Alzheimer's dementia classification using an open source dataset \cite{ds004504} available from the OpenNeuro platform \cite{OpenNeuro} to improve the reproducibility of the results.
%This dataset contains recordings from healthy individuals, patients with Alzheimer's disease and frontotemporal dementia, both progressive neurodegenerative disorders that affect elderly people, where the first accounts for 60-80\% of the cases and the latter 5–10\% of the cases \cite{casesAD}.
This dataset contains recordings from healthy individuals and patients with two distinct progressive neurodegenerative disorders: Alzheimer's disease and frontotemporal dementia. 
Both disorders affect elderly people, where Alzheimer's disease accounts for 60- 80\% of cases and frontotemporal dementia 5–10\% of cases \cite{casesAD}.

%A SIMILAR WORK RESPECTS THE DATASET
The authors of this dataset also proposed a transformer-based architecture called DICE-net in \cite{DICE-net}.
This work is significant because it combines novel attention-based techniques such as transformers, with medically informed metrics such as the Relative Band Power and the Spectral Coherence Connectivity, both of which have been shown to be important biomarkers for Alzheimer's disease \cite{DICE-net}. 
In addition, they achieved an impressive mean (unbalanced) accuracy of 83.3\% with a splitting strategy similar to Leave-One-Subject Out (LOSO) \cite{LOSO}.
%POINTS WHERE OUR WORK DIFFERS
However, their architecture has a size of 170.5 million parameters, which is significantly larger than the dataset size (88 subjects, \num{1741} 30s 50\% overlapping windows).
Moreover, they focused only on the Alzheimer's vs. healthy condition task, which is easier compared to the complete three-class classification problem addressed in this work. 
Although the philosophy behind architecture design is commendable, performance does not justify the number of trainable parameters; compared to ShallowNet, the unbalanced accuracy of ShallowNet for the same two-class task is 79.7\%, with four orders of magnitude fewer parameters than DICE-net.
It should be stressed that this and all the accuracies reported in this article should be considered as the median accuracy of all possible splits given by the Nested-Leave-N-Subjects Out (N-LNSO) training strategy, which is more conservative and provides more reliable performances than the LOSO one \cite{losoF}.

%RESULTS
This work aims to address the interpretability of ShallowNet using a different approach from that proposed by \cite{BORRA202055}, and to address the problem of architecture design in a more parsimonious way with respect to \cite{DICE-net}.
This brings several advantages, such as allowing training in smaller infrastructures and laboratories, reducing training time, and, of course, reducing overfitting.
This paper aims to fill the knowledge gap in describing the spectral properties of these types of architectures in relation to the dementia classification problem, as mentioned in the discussion section of Del Pup \& Zanola \etal~\cite{eegprepro}.
In short, it is well known that there are spectral differences between the control and Alzheimer's group, such as increased theta and decreased alpha activity in the latter \cite{Dauwels}. 
Thus, it is essential to determine whether convolutional neural networks can capture this information.

First, a theoretical description of the network is provided in \autoref{sec-networkdescription}, detailing the set of analytic functions that map input to output across different layers, while explaining in detail the underlying physical concepts.
This allows us to understand what the architecture is doing, without using gradient-based techniques, as is typical in the medical domain \cite{BORRA202055, ReviewInterp}.
Then, architectural changes are also proposed to reduce the overfitting present in ShallowNet, focusing on the learned weights and their interpretation in \autoref{sec-networkmodification}.
Overfitting will be quantified by examining the number of training epochs as well as the trend of the training and validation loss curves in \autoref{sec: overfitting}.
The overall approach aims to strike a balance between maintaining performance on the three-class classification problem at hand and reducing the number of parameters while improving interpretability.
Finally, since the median performance on this task is not great (53.7\%) and highly variable (from 34.4\% to 74.5\%), the reasons for the low median values and the high variance of the test accuracies obtained with the N-LNSO will be explained in detail for the proposed model in \autoref{sec: N-LNSO performances} and \autoref{sec: N-LNSO Variability} respectively.
The results will be presented and discussed, respectively, in \autoref{results}, \autoref{discussion}; the manuscript is then finished in \autoref{conclusions}.

\section{Methods}\label{methods}
\label{sec-methods}
%METHODS
To effectively process EEG data with deep learning models, it is important to structure the input in a manageable and meaningful way.
Let us consider an EEG recording of arbitrary length with $C$ channels.
For convolution-based DL architectures such as ShallowNet, inputting the entire recording can be challenging and inefficient.
Feeding the entire EEG recording would oversize the final dense layer, unbalancing the network, over-parameterizing the model, and increasing the overfitting risk.
Typically, the input is a time-limited segment of the EEG recording, called window, denoted as $X$.
An EEG window of length $L$ seconds, sampled at $f_s$ Hertz, is represented as a matrix $X_{C\times N}$, of dimension channels $C$ times the number of samples in the recording, which is $N=L\cdot f_s$.

%By fixing the length of the signal and the sample rate, there is a consequence in the frequency spectrum.
%\begin{itemize}
%    \item The maximum frequency that can be resolved, without aliasing is $f_{max} = f_s/2$, due to the Nyquist's Theorem \cite{SHAN}.
%    \item With a window of length $L$, the minimal frequency that can be resolved is $f_{min} = 1/L$. [TO CHECK]
%    \item Let's call $B$ the number of samples in a batch (batch size).
%    The memory required is $4B\times C\times N$ bytes, when data are stored with single precision, as required by PyTorch \cite{PyTorch}.
%\end{itemize}

\subsection{Data and Model Training}
\subsubsection{Dataset}
%DATASET
% This work focuses on the task of pathological classification with the aim of distinguishing between two forms of dementia, Alzheimer's disease and frontotemporal dementia, and healthy controls (3-class classification problem).
An important clinical application of deep learning is pathological classification, where the goal of the use case considered in this work is to distinguish between two forms of dementia, Alzheimer's disease and frontotemporal dementia, from the healthy controls in a three-class classification problem.
The dataset used in this study is publicly available on OpenNeuro under the title ``A dataset of EEG recordings from: Alzheimer's disease, Frontotemporal dementia and Healthy subjects"~\cite{ds004504}.
It contains resting-state EEG recordings with eyes closed (duration $13.4 \pm 2.3$ minutes) from 36 Alzheimer's patients, 23 subjects diagnosed with frontotemporal dementia, and 29 age-matched healthy controls (respectively $66.4\pm 7.9$ years, $63.7\pm8.2$ years, and $67.9\pm5.4$ years).
The EEG recordings have nineteen channels ($C=19$) distributed according to the international 10-20 system \cite{Jasper}.

\subsubsection{Preprocessing}
%PREPROCESSING
In Del Pup \& Zanola \etal~\cite{eegprepro}, the authors investigated how EEG preprocessing influences performance in six different tasks. 
They found that a simple filtering operation consistently outperforms more complex preprocessing strategies in most cases, making it a more effective and efficient choice for deep learning applications.
Following this guideline, the recordings were downsampled to $f_s=125$Hz and the continuum (DC) component of each channel was removed; the recordings were also filtered between [1,45]Hz with a finite impulse response (FIR) filter and referenced to the mean.
The whole preprocessing was done with a \textit{BIDSAlign}, a MATLAB\textsuperscript{\textregistered} based library allowing full preprocessing and standardization of EEG data~\footnote{\href{https://github.com/MedMaxLab/BIDSAlign}{MedMaxLab/BIDSAlign}} (see Zanola \etal~\cite{Zanola_2024}).

\subsubsection{Training strategy}
%TRAINING
The EEG recordings were divided into windows of length $L=4$s, with zero overlap between them, and all windows coming from one subject are assigned to the training, validation or test set, in a subject-based split strategy.
The final portion of the recording was discarded if it was not long enough to generate an additional sample.
The EEG windows have been z-scored channel-wise, before being given to the model.
In addition, the Nested-Leave-N-Subjects Out (N-LNSO) validation strategy was used to have the best possible estimate of the real performance of the network on unseen data, taking into account subject variability, as in \cite{eegprepro}.
The N-LNSO strategy is built on ten outer folds and five inner ones, for a total of 50 different splits.
In each split, $N$ subjects are isolated for testing (outer split), while the rest are divided into training and validation (inner split). 
Splits are generated by cycling subject assignments across the three sets.
Since it is inconvenient to visualize all the weights learned by all the models trained in the procedure, some results are shown for either the worst split or for the worst and best splits.
The assignment of subjects to training, validation and test sets was done in a stratified way; the proportions between training, validation, and test sets are respectively 72\% $(\text{CI}_{95\%}=[70,75]\%)$, 18\% $(\text{CI}_{95\%}=[16,20]\%)$ and 10\% $(\text{CI}_{95\%}=[9,11]\%)$.

\subsubsection{Hyper-parameters}
%HYPER-PARAMETERS
All trainings were conducted using a custom seed randomly chosen (83136297) to improve the reproducibility of the results. 
Each of the 50 models (one for each split) was initialized using the default settings of PyTorch \cite{PyTorch}, which are specific to each type of layer, ensuring that the initialization of the network was always the same.
The models were trained using Adam optimizer $(\beta_1 = 0.9,~\beta_2 = 0.999$, without weight decay), batch size $B=64$, cross entropy as loss function, and an exponential learning rate scheduler $(\gamma = 0.995, ~l_0=5\cdot10^{-5})$.
The maximum number of epochs was set at \num{1000}, coupled with an Early Stopping \cite{earlystopping} with patience of 15 epochs (monitoring the validation loss) and with a minimal improvement on the monitored quantity of $\Delta_{min}=10^{-4}$.

\subsubsection{Software and Hardware Configuration}
\textit{SelfEEG} v0.2.0 \footnote{\href{https://github.com/MedMaxLab/selfEEG}{MedMaxLab/selfEEG}} \cite{selfeeg}, a Python library based on PyTorch, was used to perform experiments and model trainings.
The experiments were carried out on an NVIDIA\textsuperscript{\textregistered} A30 GPU device.

\subsection{Architectural baseline: ShallowNet}
\label{sec-networkdescription}
The description of ShallowNet follows the one presented in the original paper \cite{shallow}.
A tensor (multidimensional array) notation accompanies the description to be consistent also with conventions in deep learning practice.
The PyTorch description of the architecture can be found in the supplementary material section 1.

\subsubsection{First Layer}
The first layer in ShallowNet is a 2D convolutional one of dimension $(\mathcal{F}_1,\mathcal{K}_1,H_1,W_1)=(40,1,1,25)$, with $\mathcal{F}_1$ filters, respectively, with $\mathcal{K}_1$ horizontal kernels, moved with a stride $S_1$ equal to one; the input signal does not have padding.
The input EEG signal is unsqueezed to match the expected 2D format.
This layer takes as input a tensor $X$ of dimensions $(B,\mathcal{F}_0,C,N_0)=(64,1,19,500)$, and outputs a tensor of dimensions $(B,\mathcal{F}_1,C,N_1)=(64,40,19,476)$.
The analytical function of the first layer is
\begin{equation}
    \tilde{X}^{urp}_{ji}=b^p + \sum_{t=0}^{W_1-1}{k^{p}_{t}X^{ur}_{j\,i+t}}
    \label{equ: conv1}
\end{equation}
where the following notation is used for the rest of the work.
\begin{itemize}
    \item[--] $j,~i$ runs respectively over the $C$ channels and $N_1$ time steps;
    \item[--] $r$ runs over the $B$ samples in the $u^{\text{th}}$ batch; $u$ runs over the $U$ batches of the training.
    \item[--] $p$ runs over the $\mathcal{F}_1$ filters;
    \item[--] $b^{p},~k^{p}$ are the bias and kernel of the $p^{\text{th}}$ filter; the latter convolved with the input gives the $p^{\text{th}}$ feature map;
    \item[--] $X^{ur}$ is the $r^{\text{th}}$ input window in the $u^{\text{th}}$ batch;
    \item[--] $\tilde{X}^{urp}$ is the $p^{\text{th}}$ feature map, obtained from $X^{ur}$. 
\end{itemize}
The result of the first layer is $\mathcal{F}_1$ filtered versions of the input EEG window, of length $N_1=476$
\footnote{$N_1=\lfloor\frac{N_{0}-W_1}{S_1}+1 \rfloor$}.
Considering the direction in which the kernel operates, as mentioned in other works \cite{BORRA202055, Wang} they act as temporal digital filters.
In detail, the kernels $\mathcal{K}_1$ act as type 2 finite impulse response (FIR) filters, so the Bode plots can be displayed to see the frequency response.
As can be seen in section 2 of the supplementary material, the kernels attempt to mimic the behavior of band-pass/band-stop filters, but the magnitudes rarely go below -20dB which is, as a rule of thumb, considered a sufficient attenuation.

The frequency responses of the kernels are not explicitly learned during training; rather, they primarily reflect the responses of randomly initialized filters.
To prove this, for both the best and the worst split, the distribution of weight variation can be calculated, and shown in section 2 of the supplementary material.
By defining $\Delta_{w\%} = \frac{w-w_{init}}{|w|}\cdot100$, where $w$ is the weight after training and $w_{init}$ is its value at the initialization, it can be seen that $\Delta_{w\%}$ range is small, with a distribution highly peaked in zero.
This means that the frequency responses of the filters in the first layer before and after the training are almost the same.
Consequently, even after training, ShallowNet has random filters in the first layer.
This suggests that since the median performance of ShallowNet is the highest of all its variants (see \autoref{fig: architecture_var}), it is unlikely that the features learned in the first layer are the ones contributing more to solve the task.

\subsubsection{Second Layer}
The second layer in ShallowNet is a 2D convolutional one of dimension $(\mathcal{F}_2,\mathcal{K}_2,H_2,W_2)=(40,40,19,1)$, with $\mathcal{F}_2$ filters, respectively with $\mathcal{K}_2$ vertical kernels, moved with a stride $S_2$ equal to one; the input signal has no padding.
Thus the kernels of the second layer, as mentioned by other works \cite{shallow, BORRA202055}, act as spatial filters, i.e they weight electrodes differently, possibly changing their polarity using negative values.
This layer takes as input a tensor $\tilde{X}$ of dimensions $(B,\mathcal{F}_1,C,N_1)=(64,40,19,476)$, and outputs a tensor of dimensions $(B,\mathcal{F}_2,1,N_2=N_1)=(64,40,1,476)$.
The analytical function of the second layer is
\begin{equation}
    \hat{X}^{urs}_{i} = b^s + \sum_{p=1}^{\mathcal{F}_1}\sum_{c=0}^{H_2-1}{k^{sp}_{c}\tilde{X}^{urp}_{ci}}
    % \hat{X}^{rs}_{i} = b^{s} + \sum_{p=1}^{\mathcal{F}_1}b^p\sum_{c=0}^{H_2-1}{k^{sp}_{c}} + \sum_{p=1}^{\mathcal{F}_1}\sum_{c=0}^{H_2-1}k^{sp}_{c}\sum_{a=0}^{W_1-1}{k^p_{a}X^{r}_{ji+a}}
    \label{equ: conv2}
\end{equation}
where
\begin{itemize}
    \item[--] $s$ runs over the $\mathcal{F}_2$ filters;
    % \item[--] $k^{sp}$ is the $s^{\text{th}}$ kernel, applied to the $p^{\text{th}}$ input feature map;
    % \item[--] $b^{s}$ is the $s^{\text{th}}$ bias, of the $s^{\text{th}}$ filter;
    \item[--] $b^{s},~k^{sp}$ are the bias and kernel of the $s^{\text{th}}$ filter; the latter applied to the $p^{\text{th}}$ input feature map;
    \item[--] $\hat{X}^{urs}$ is the $s^{\text{th}}$ feature map, for the $r^{\text{th}}$ sample in the $u^{\text{th}}$ batch.
\end{itemize}
The result of the second layer is $\mathcal{F}_2$ time-filtered and spatial-weighted versions of the input EEG window of length $N_1$.
In fact, as can be seen in \autoref{equ: conv2}, a feature map $\hat{X}^{rs}$, is a weighted sum over all the electrodes, and over all the feature maps from the first layer.

\subsubsection{Batch Normalization}
\label{sec: batch}
The third layer in ShallowNet targets batch normalization \cite{batchnorm}, a regularization technique found to be useful and effective in training deep learning models.
The layer performs a batch- and feature-wise normalization, meaning that it takes as input a tensor $X$ of dimensions $(B,\mathcal{F}_2,1,N_2=N_1)=(64,40,1,476)$, and outputs a tensor of equal dimensions.
The output for each of the $\mathcal{F}_2$ feature map is 
\begin{equation}
     \bar{X}^{urs} = \frac{\hat{X}^{urs} - [(1-m) \mu^{us} + (m)\mu^{u-1\,s}]}{\sqrt{[(1-m)\text{Var}^{us} + (m)\text{Var}^{u-1\,s}]}}\gamma + \beta
     \label{equ: batch}
\end{equation}
where $m$ indicates the momentum, $\mu$ the mean and $\text{Var}$ the variance \footnote{$\quad
    \mu^{us} = \hat{X}^{urs} - \frac{1}{BN_1}\sum_{r=1}^{B}\sum_{i=1}^{N_1}\hat{X}^{urs}_{i},\quad \text{Var}^{us} = \frac{\sum_{r=1}^{B}\sum_{i=1}^{N_1}(\hat{X}^{urs}_{i}-\mu^{us})^2}{N_1-1}$}.
ShallowNet employs affine transformations in the batch normalization layer, making the parameters $\gamma$ and $\beta$ learnable.
The default values of the hyper-parameters used are $\gamma=1, \beta=0, m=0.1$; when $m=0$, \autoref{equ: batch} becomes the z-score.

\subsubsection{Average Pooling}
After batch normalization, ShallowNet performs two algebraic operations to extract useful temporal information from the features given by the previous layer.
Indeed, ShallowNet does the square of $\bar{X}^{urs}$; this has profound implications for interpretation.
Until now, the feature maps are just spatially weighted versions of the inputs filtered in time, but still related to the EEG time signal; in other words, the physical unit of these feature maps is still expressed in $\mu V$.
The squared operation gives a positive bounded signal, which is strictly related to the energy of the feature maps $\bar{X}^{urs}$.
When this operation is coupled with averaging, the power of the signal is calculated.

The average pooling used by ShallowNet has a length of $W_3=75$, with a stride of $S_3 = 15$, which corresponds to a time window of $W_3/f_s = 600ms$, with a time resolution of $S_3/f_s=120ms$.
This allows to calculate the power of the feature map $\bar{X}^{urs}$ at different time points.
As shown in Cui \etal \cite{TowardBestPractice}, DL architectures are able to identify ocular artifacts such as vertical and horizontal blinks and use these physiological signals to perform the required task. 
These artifacts are, in fact, powerful low-bursts localized in time, which can be detected with a suitable time resolution.
The number of feature maps obtained in the output is $N_3=27$ \footnote{$N_3=\lfloor\frac{N_2-W_3}{S_3}+1 \rfloor$}.
The analytical formulation of this layer 
is
\begin{equation}
    P^{urs}_l = \frac{1}{W_3} \sum_{t=0}^{W_3-1}|\bar{X}^{urs}_{t+(l-1)S_3}|^2\quad \text{where}~l\in\{1,\dots,N_3\}
    \label{equ: avgpool}
\end{equation}

\subsubsection{Non Linear Activation}
The next step is a non-linear transformation of the previous power $P^{urs}_l$.
The original paper used the natural logarithm; however with a general formulation, $\hat{P}^{urs}_{l}$ can measure the power in decibel (if $c_1=c_2=10$); the constant factor will be denoted as $K$.
\begin{equation}
    \hat{P}^{urs}_{l} = c_1log_{c_2}(P^{urs}_{l})=\frac{c_1}{log_{e}(c_2)}log_{e}(P^{urs}_{l})
\end{equation}
The value of power $P^{urs}_l$ is limited in the interval $[10^{-7},10^4]$ for numerical stability.

\subsubsection{Dense Layer}
A flattening operation must be performed before applying the final dense layer.
Here is also where ShallowNet applies dropout \cite{dropout}, another well-known regularization technique.
During training, the flatten layer activations are point-wise multiplied by the dropout mask $d$ $(p=0.2)$; during testing, dropout is inactive $(p=0)$.
The length of the flatten layer is $\mathcal{N}_{in}=\mathcal{F}_2\cdot N_3=$\num{1080}, which is the number of input neurons in the dense layer, while the number of output neurons is equal to the classes present in the problem.
$w_{cn}$ will denote the weight between an input neuron $n\in \{1,\dots,\mathcal{N}_{in}\}$ and the output class $c\in \{1, \dots, \mathcal{L}\}$.

\subsubsection{SoftMax Activation}
\label{softmax-activ}
The activation of each output neuron is determined by the weighted sum of the previous dense layer's inputs and biases. 
The softmax function then normalizes these activations to produce probability values for classification $p_c^{ur}$.
The class is predicted by taking the maximum between the $\mathcal{L}$ classes' probabilities; this can be equivalently done over the logarithm of them.
Considering that the dense weights can be either positive ($\mathcal{N}_{+}$) or negative ($\mathcal{N}_{-}$), the formula above can be rewritten as
\begin{equation}
    log_e(p_c^{ur})
    \propto \frac{b_c}{K} + \sum_{n=1}^{\mathcal{N}_+} |w_{cn}|d_{n}log_{e}{P^{urs}_{l}}- \sum_{n=1}^{\mathcal{N}_-} |w_{cn}|d_{n}log_{e}{P^{urs}_{l}}\\
    \label{equ: class_pred}
\end{equation}
or
\begin{equation}
    log_e(p_c^{ur})
    \propto\frac{b_c}{K} + log_{e}\biggl[\frac{\prod_{n=1}^{\mathcal{N}_+} {(P^{urs}_{l}})^{\,^{|w_{cn}|d_{n}}}}{\prod_{n=1}^{\mathcal{N}_-} {(P^{urs}_{l})}^{\,^{|w_{cn}|d_{n}}}}\biggr]
    \label{equ: class_pred2}
\end{equation}
Equations \ref{equ: class_pred} and \ref{equ: class_pred2} clearly state that for an EEG window, the most likely class is the one in which the sum of the powers $P^{urs}_l$, weighted by the positive weights $\mathcal{N}_{+}$, is greater respect the sum of the powers $P^{urs}_l$, weighted by the negative weights $\mathcal{N}_{-}$.
Their derivation can be found in the supplementary material Section 3.

Due to Parseval's theorem, the total energy of a signal can be calculated by summing either the power per sample in the time domain or the spectral power along the frequency domain \cite{parseval}.
This supports the claims of ShallowNet's creators \cite{shallow} that the architecture learns how to use (and extract) the frequency bands relevant to the task at hand. 

\subsection{Enhancing ShallowNet: xEEGNet}
\label{sec-networkmodification}
Building on the theoretical insights outlined above, the following sections detail the modifications made to develop xEEGNet, the enhanced model proposed in this study.
It is structured as follows:
\begin{itemize}
    \item[--] the first layer is frozen, with $\mathcal{F}_1^{'}=7$ pre-initialized kernels of length $W_1=125$ and no bias;
    \item[--] the second layer is a depthwise convolution, with depth equal to one and no bias;
    \item[--] the batch normalization has affine transformation, with momentum $m=0.1$;
    \item[--] the squared activation precedes global average pooling, followed by a non-linear activation in decibels, given by $10\log_{10}(\cdot)$;
    \item[--] dropout $p=0.2$ is applied to the dense layer, which is without bias.
\end{itemize}
The size of the network is now only a function of the number of initial filters and classes, the latter being fixed a priori.
The number of trainable parameters is $\mathcal{F}_1^{'}(C+\mathcal{L}+2)=168$.
The PyTorch description of the architecture can be found in the supplementary material section 1.

\subsubsection{First Layer}
\label{sec-firstlayermed}
xEEGNet improves the first layer of ShallowNet by including pre-designed biological meaningful filters targeting specific EEG bands.
As shown in the supplementary material section 2, the kernels in the first layer acts as type 2 FIR filters, which provide small attenuation with wide and poorly defined transition bands.
% Using Harris' approximation \cite{FH_ruleofthumb}% (page 752, equation B.39, \ref{equ: harris}) 
% (page 752, equation B.39) to estimate the filter order, with a desired attenuation of $A = 20$dB and a transition band of $\Delta f = 1$Hz, it was determined that the filter kernels should have a length approximately of 113; in this work, a length of 125 will be used.
Using Harris' approximation \cite{FH_ruleofthumb} (Equ. B.39, p. 752)\footnote{$W_1 \approx \frac{A}{22dB}\frac{f_s}{\Delta f}=\frac{20 dB\cdot125 Hz}{22 dB\cdot1 Hz}=114$}, the estimated filter order for $20$dB attenuation and $1$Hz of transition band is approximately 114; here, a length of 125 is used.

% \begin{equation}
%     W_1 \approx \frac{A}{22dB}\frac{f_s}{\Delta f}
%     \label{equ: harris}
% \end{equation}
%It is worth to underline, that longer filters turns out in shorter feature maps $\mathcal{FM}_1$, since padding is not used.
To obtain biologically meaningful filters, they are designed to pass or eliminate specific EEG bands, which have the following frequency range, expressed in Hertz: (delta) $\delta = [1,4]$, (theta) $\theta = (4,8]$, (alpha) $\alpha = (8,12]$ \cite{Mesulam}; (beta-1) $\beta_1 = (12,16]$,  (beta-2) $\beta_2 = (16,20]$,  (beta-3) $\beta_3 = (20,28]$ \cite{rangaswamy}; finally (gamma) $\gamma = (28,45]$ with the left uncover frequency range.
Considering all band combinations, 127 kernels are needed, but the number can be progressively reduced to a minimum of seven, one per EEG band ($\mathcal{F}_1^{'}=7)$.
As shown in the supplementary material section 2 respect the random initialization, ShallowNet's weights are either no or mildly modified during the training.
Consequently the first layer can be frozen in order to decrease the number of trainable parameters to zero, while before was $\mathcal{F}_1 (\mathcal{K}_1\cdot W_1 + 1) = 1040$.
Once the filters were pre-initialized and frozen, the bias was avoided as well.
%The \textit{firwin} function from the Python library \textit{Scipy} \cite{Scipy} was used for filters design.

\subsubsection{Second Layer}
The second layer of ShallowNet takes each feature map in the first layer, multiplies it by a specific vertical kernel, and then averages over the number of feature maps $p$.
This operation is performed $s$ times, with different filters.
To increase the interpretability, inspired by these works \cite{eegnet, BORRA202055}, the depthwise convolutional layer with depth equal to one is used.
This setup allows each feature map from the first layer to be multiplied by a single, dedicated kernel in the second layer, producing a corresponding feature map in the second layer.
The number of filters in this layer is seven, i.e the EEG bands defined in \autoref{sec-firstlayermed}; in this layer, the use of bias was intentionally omitted.
This reduces the number of parameters from $\mathcal{F}_2(\mathcal{K}_2\cdot H_2+1) = $ \num{30440} to $\mathcal{F}_1^{'}\cdot H_2 = 133$.
As written in \cite{BORRA202055} this layer makes the interpretation of the features learned easier, since it associates one EEG band to one scalp topography.

\subsubsection{Average Pooling and Dense Layer}
ShallowNet uses average pooling to estimate feature map power over time, preserving temporal information for classification. 
This is useful for Event-Related Potential tasks, where the localization in time of the signal is important, but less relevant for pathology classification, where class labels are time-invariants.
Global Average Pooling \cite{globavg} can then be used, since it naturally introduces translational invariance in space \cite{globavgtimeinv} (here, in time).
Given the clear interpretability provided by \autoref{equ: class_pred} and \autoref{equ: class_pred2} with respect to class prediction, the bias term in the dense layer can be safely omitted.
Omitting the bias term simplifies the correspondence between the logits and the flattened activation values, while also reducing the number of trainable parameters in the dense layer from $(\mathcal{F}_2\cdot\mathcal{F}_3+1) \mathcal{L} =$ \num{3243} to $\mathcal{F}_1^{'}\cdot\mathcal{L} = 21$.

Evidence that the model is capable of extracting the frequency content of the signal is shown in \autoref{fig: PSD_corr}.
In the figure, the flatten activations for the seven EEG bands $\mathcal{F}^{n}$, are correlated respectively with the average power bands (over channels) derived from the EEG windows of the test set $\mathcal{F}_X^{n}$.
The average power bands (in dB) of an EEG window in the $n^{\text{th}}$ EEG band defined by the intervals $[f_L,f_H]$ is calculated as:
\begin{equation}
    %\mathcal{F}_X^{n} = 10\log_{10}\biggl( \frac{1}{C}\sum_{c=1}^C \frac{1}{f_H-f_L}\int_{f_L}^{f_H}PSD_c(X)~df \biggr)
    \mathcal{F}_X^{n} = 10\log_{10}\biggl( \frac{1}{C}\sum_{c=1}^C \int_{f_L}^{f_H}PSD^X_c(f)~df \biggr)
    \label{equ: Fx}
\end{equation}
Using the Welch’s method, the power spectral density $(PSD^X_c)$ is estimated for each EEG window $X$ and channel $c$, with 50\% overlapping segments of length $N/2$.

\subsection{Evaluating xEEGNet for Dementia Classification}
\subsubsection{Overfitting}
\label{sec: overfitting}
% Overfitting is the statistical phenomenon where a model reduces the training loss, while validation loss increases.
% As the name suggests, it is thought of as an adaptation (fitting) of the model to the training data, to the extent that is actually a misfit and the generalisation does not hold for the validation data.
Overfitting is a statistical phenomenon in which a model decreases training loss while validation loss increases. 
As the name suggests, it refers to the model fitting the training data too closely, which leads to poor generalization and reduced performance on validation data.

% overfitting STUDIED AS: CORRELATION BETWEEN LOSSES
Overfitting can be analyzed through the learning dynamics of a single model, with validation error evolving over training epochs. 
It often follows a proper learning phase, in which both training and validation losses initially decrease. 
Eventually, they become negatively correlated, with further training improving the fit to training data while worsening the validation performance \cite{overfitpaper}.
The correlation between training and validation losses between models is shown in \autoref{fig: curve_loss} (right panels). 
In this context, overfitting arises when the model learns both signal and noise, with the latter failing to generalize to unseen data, by definition.
To mitigate this, Early Stopping \cite{earlystopping} is commonly used, halting training once validation loss ceases to improve significantly.
% overfitting STUDIED AS: NUMBER OF EPOCHS 
This technique also serves as an indirect indicator of overfitting: if training halts well before reaching the maximum number of allowed epochs, it suggests that the model quickly converges to a local minimum in the validation loss and may have limited capacity to discover deeper, more optimal minima.
The training epochs across models are analyzed in \autoref{fig: architecture_var} Panel B.

% overfitting STUDIED AS: MODEL SIZE
An alternative perspective on overfitting focuses on model families with differing parameter counts.
For a given dataset and fixed signal-to-noise ratio (potentially achieved through preprocessing), increasing the number of parameters generally leads to a monotonic decrease in the final or best training loss.
%In parallel, there is an association between the size of the training data and the parameter budget.
Unsurprisingly, large models trained on small datasets often achieve very low training loss but high validation loss, exhibiting classic overfitting symptoms (\autoref{fig: curve_loss}, central panels). For a given dataset, validation loss initially decreases with model size, reaching a minimum at a certain parameter range before rising again as the model grows \cite{overfitpaper}.

Overall, selecting an appropriate model size and training procedure is essential to effectively match the characteristics of the given dataset.
With EEG data, very shallow and very deep models show approximately the same performance on validation and test data (see \autoref{fig: architecture_var}, Panel A).
Additionally, models trained with Early Stopping undergo only a few training epochs (\autoref{fig: architecture_var}, Panel B), suggesting that some may be modeling noise.
In this context, noise encompasses not only random measurement noise but also structured or deterministic patterns that, while present in the training data, are uncorrelated or irrelevant to the task when generalizing to unseen data.
Thus, monitoring validation-training loss correlation, its dynamics, and the duration of effective learning is essential for proper learning.

\subsubsection{Neasted-Leave-N-Subjects Out Performances}
\label{sec: N-LNSO performances}
The main aim of this section is to investigate the reasons behind the performance of the proposed model xEEGNet and to compare them with other machine learning and deep learning approaches.
%is relatively low (53.7\%).%, while the one of \autoref{sec: N-LNSO Variability} is to investigate why the performances have a high variability (from 34.4\% to 74.5\%).
The theoretical discussion in \autoref{softmax-activ} and the computational results in \autoref{sec: denselayer} show that xEEGNet uses the power spectrum of the signal to predict the pathology.
While in the proposed architecture the interpretation is immediately understandable, in ShallowNet the function of the different layers was not clear.
Now that xEEGNet has been made a ``white-box'' architecture, it is interesting to understand if a simple Multinomial Logistic Regression Model (MLRM) \cite{MLRM} could have the same performance as a neural network.
MLRM is a model taken from statistics, where the alternative hypothesis is that the model gives significantly better results than random chance.
This comparison thus, shows how xEEGNet serves as a bridge between deep neural networks and statistical models.
In addition, a hybrid model, `shnMLRM', is introduced, combining xEEGNet with a dense layer replaced by an MLRM.
In this way, at least three comparisons can be made:
\begin{itemize}
    \item[--] MRLM vs. ShallowNet: the comparison will show the differences in performances between a conventional statistical technique and a deep neural network;
    \item[--] MLRM vs. `shnMLRM': the comparison will show the effect of extracting the spectral content of each band (using xEEGNet encoder) respect calculating it explicitly from the data;
    \item[--] xEEGNet vs. `shnMLRM': the comparison will show the effect of substituting the dense layer with a MLRM model.
\end{itemize}
For each of the four models the same N-LNSO training strategy is used; notice that MLRM does not require a validation set.
%The \textit{MNLogit} function from the Python library \textit{statsmodels} \cite{statmodels} was used to train the MLRM model.
%The Newton-conjugate gradient method was used as optimizer for the model fit.
The comparison between these four models is reported in the result section in \autoref{fig: MLR}.

\subsubsection{Impact of Data Partitioning on Performance Variability}
\label{sec: N-LNSO Variability}
The final question is why the proposed model's performance varies significantly, ranging from 34.4\% to 74.5\%. 
This variability is mainly driven by the selection of triplet training, validation, and test sets, which form the core concept of N-LNSO as described in \cite{eegprepro}.
% In this space, using a Gaussian Kernel Density Estimate (KDE), we can find the one-sigma 3D surface containing 68\% of the data, which can be considered ``the bulk'' of the training set distribution.
% Thus, it is possible to count how many samples of the test set are within the ``bulk" of the training set; this defines a quantity that is the ``overlap'' between the training and test sets.
% The function \textit{gaussian\_kde} from the \textit{SciPy} \cite{Scipy} Python library was used to find the one-sigma 3D surface of the training set.
To demonstrate this, the logits of the three output neurons are used as class coordinates. 
High performance should occur when classes are well separated in logit space, either due to inherent separability in the input data or due to learning by the network. 
However, separability between classes alone cannot explain performance, so it is crucial to account for separability between sets.
To address this, embedding points (EEG windows) are grouped by label (Alzheimer's disease, frontotemporal dementia, control) and set (training, validation, test), forming 9 groups and 36 possible pairwise combinations.
The separability definition adopted here, assumes that the cloud of points of each of these 36 groups has a spherical shape, a relative mild assumption; in details, between two groups $\mathbb{I}$ and $\mathbb{J}$ the separability is defined as
\begin{equation}
    o_{ij} = \frac{d_{ij}}{r_i + r_j}
    \label{equ: separability}
\end{equation}
Here $d_{ij}$ is the euclidean distance between the barycenter of the two sets, and $r_i,~r_j$ represent the radii, calculated as the average distance of points in each set from their respective barycenters.
A visual representation of this can be found in supplementary material, section 4.

If $d_{ij}>(r_i+r_j)$, the groups are well separated $(o_{ij}>1)$, meaning less overlap.
Labels AD, FTD, CTL denote the three classes, while $tr,v,ts$ represent training, validation and test sets.
A combination is written as $label_{\text{set}}$ (e.g., $\text{AD}_{tr}$, for Alzheimer's in training).
Separability between two groups is noted as  $\text{group}_1~-~\text{group}_2$ (e.g $\text{AD}_{tr}~-~\text{CTL}_{ts}$ for Alzheimer's in training vs. control in test).
This methodology calculates separability for all possible pairs and correlates it with weighted accuracy using an ordinary least squares (OLS) regression model. 
% The function \textit{regression.linear\_model.OLS} from the \textit{statsmodels} \cite{statmodels} Python library was used to perform this regression analysis.
Multicollinearity can arise when multiple independent variables are used, meaning that some variables depend on others \cite{multicoll}.
To reduce features, Pearson’s correlation between weighted accuracy and separability values is computed, followed by Holm's correction \cite{holm} for multiple comparisons. 
Only significantly correlated variables ($\alpha=0.05$) are retained.
Feature selection is performed using the Forward-Backward Feature Selection algorithm \cite{FBFS1,FBFS2}, which iteratively adds significant features (based on p-values) and removes those that do not improve the model.
Once selected, the Variance Inflation Factor (VIF) is calculated to check for multicollinearity, and features with $\text{VIF}>5$ are removed \cite{VIF}.
The final optimal feature set is used to fit the regression model, and results for the xEEGNet model are reported in \autoref{sec: N-LNSO Variability res}.

\section{Results}\label{results}
\subsection{Enhancing ShallowNet: xEEGNet}
From the analytical functions described in \autoref{sec-networkdescription}, several suggestions have been proposed in \autoref{sec-networkmodification} in order to increase the interpretability of the network. 
The progressive application of the various proposed changes gives an ensemble of different models that can be evaluated; these models are described in \autoref{tab:weights}.

\setlength{\tabcolsep}{2.5pt}
\begin{table}[h]
    \centering
    \footnotesize
\begin{tabular}{|c|c|c|c|c|c|c|c|c|c|c|c|c|c|}
     \hline
          & \multicolumn{5}{c|}{\textbf{First Layer}} & \multicolumn{3}{c|}{\textbf{Second Layer}} 
          & \multicolumn{1}{c|}{\textbf{}} & \multicolumn{2}{c|}{\textbf{Pooling}} & \multicolumn{1}{c|}{\textbf{Dense}}  \\ 
     \cline{2-9}
     \cline{11-13}
      \textbf{Model} & \textbf{$F_1$} & \textbf{$K_1$} & \textbf{Bias} & \textbf{Init} & \textbf{Frozen} & \textbf{$F_2$} & \textbf{Bias} & \textbf{Depthwise} & \textbf{Log} & \textbf{Type} & \textbf{Pool} & \textbf{Bias} \\
      \hline
      ShallowNet & 40 & 25 & \checkmark  &  &  & 40 & \checkmark &  & $\log_e$ & Average & 75 &  \checkmark\\
      \hline
      shn1 & 127 & 25 & \checkmark  &  &  & 40 & \checkmark &  & $\log_e$ & Average & 75 & \checkmark \\  
      \hline
      shn2 & 127 & 125 &\checkmark   &  &  & 40 & \checkmark &  & $\log_e$ & Average & 75 & \checkmark \\
      \hline
      shn3 & 127 & 125 &\checkmark   & \checkmark &  & 40 & \checkmark &  & $\log_e$ & Average & 75 & \checkmark \\
      \hline
      shn4 & 127 & 125 &\checkmark   & \checkmark & \checkmark & 40 & \checkmark &  & $\log_e$ & Average & 75 &  \checkmark\\
      \hline
      shn5 & 127 & 125 &\checkmark   & \checkmark & \checkmark & 127 & \checkmark & \checkmark & $\log_e$ & Average & 75 & \checkmark \\
      \hline
      shn$6_{127}$ & 127 & 125 & \checkmark  & \checkmark & \checkmark & 127 & \checkmark & \checkmark & $\log_e$ & Global &  & \checkmark \\
      \hline
      shn$6_{126}$ & 126 & 125 & \checkmark  & \checkmark & \checkmark & 126 & \checkmark & \checkmark & $\log_e$ & Global &  & \checkmark \\
      \hline
      shn$6_{119}$ & 119 & 125 &\checkmark   & \checkmark & \checkmark & 119 & \checkmark & \checkmark & $\log_e$ & Global &  & \checkmark \\
      \hline
      shn$6_{98}$ & 98 & 125 &\checkmark   & \checkmark & \checkmark & 98 & \checkmark & \checkmark & $\log_e$ & Global &  & \checkmark \\
      \hline
      shn$6_{63}$ & 63 & 125 & \checkmark  & \checkmark & \checkmark & 63 & \checkmark & \checkmark & $\log_e$ & Global &  & \checkmark \\
      \hline
      shn$6_{28}$ & 28 & 125 & \checkmark  & \checkmark & \checkmark & 28 & \checkmark & \checkmark & $\log_e$ & Global &  & \checkmark \\
      \hline
      shn$6_{7}$ & 7 & 125 &\checkmark   & \checkmark & \checkmark & 7 & \checkmark & \checkmark & $\log_e$ & Global &  & \checkmark \\
      \hline
      shn7 & 7 & 125 &  & \checkmark & \checkmark & 7 &  & \checkmark & $\log_{e}$ & Global &  &  \\
      \hline
      xEEGNet & 7 & 125 &  & \checkmark & \checkmark & 7 &  & \checkmark & $10\log_{10}$ & Global &  &  \\
      \hline
\end{tabular}
    \caption{Architecture variation. 
    This table show the models tested and the various differences.
    On the top there is the original model ShallowNet, while on the bottom xEEGNet, the proposed model.
    From model `shn1' to model `shn7' there are architectures in which only one parameter is progressively changed (except in `shn5', because depthwise layer requires $\mathcal{F}_1$ to be equal to $\mathcal{F}_2$).
    Every model has the batch-normalization with trainable affine transformation, the squared activation and dropout ($p=0.2$).
    `Init' stays for initialization.}
    \label{tab:weights}
\end{table}

\begin{figure}[h!]
    \centering
    \includegraphics[width=1\linewidth]{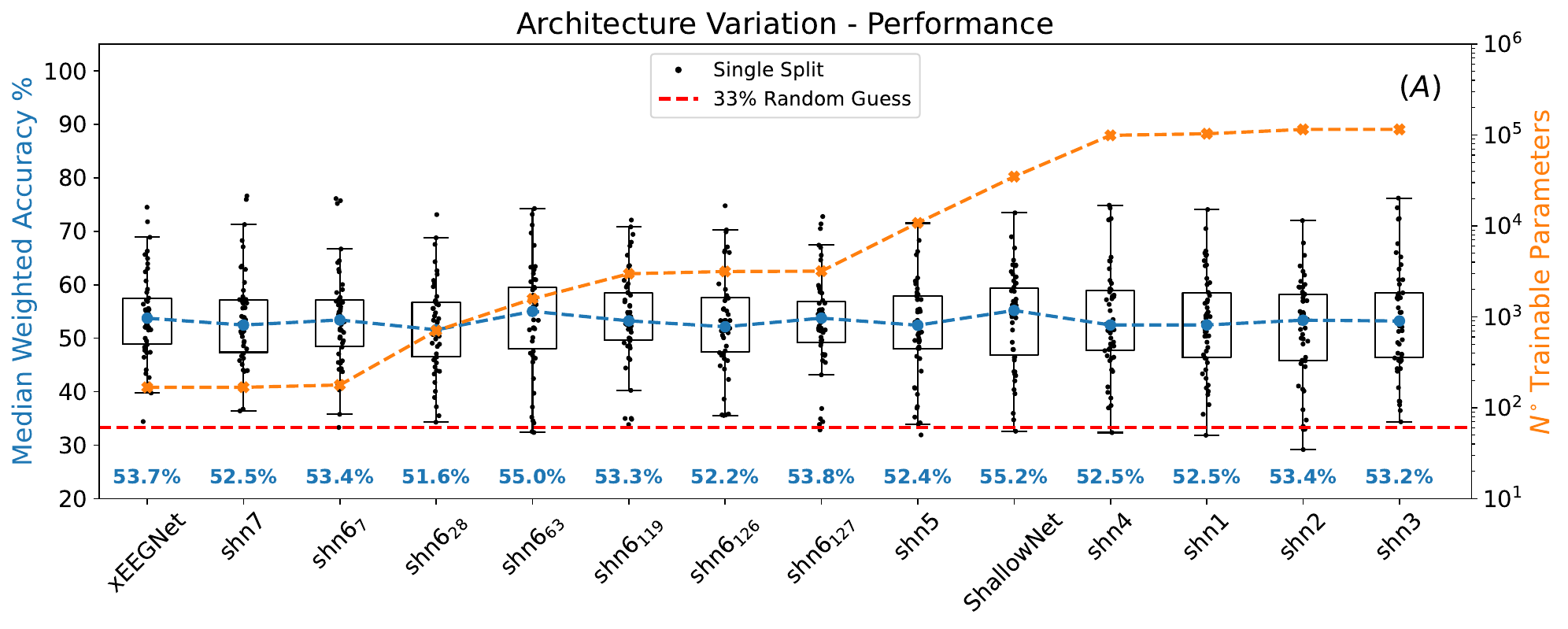}
    \includegraphics[width=1\linewidth]{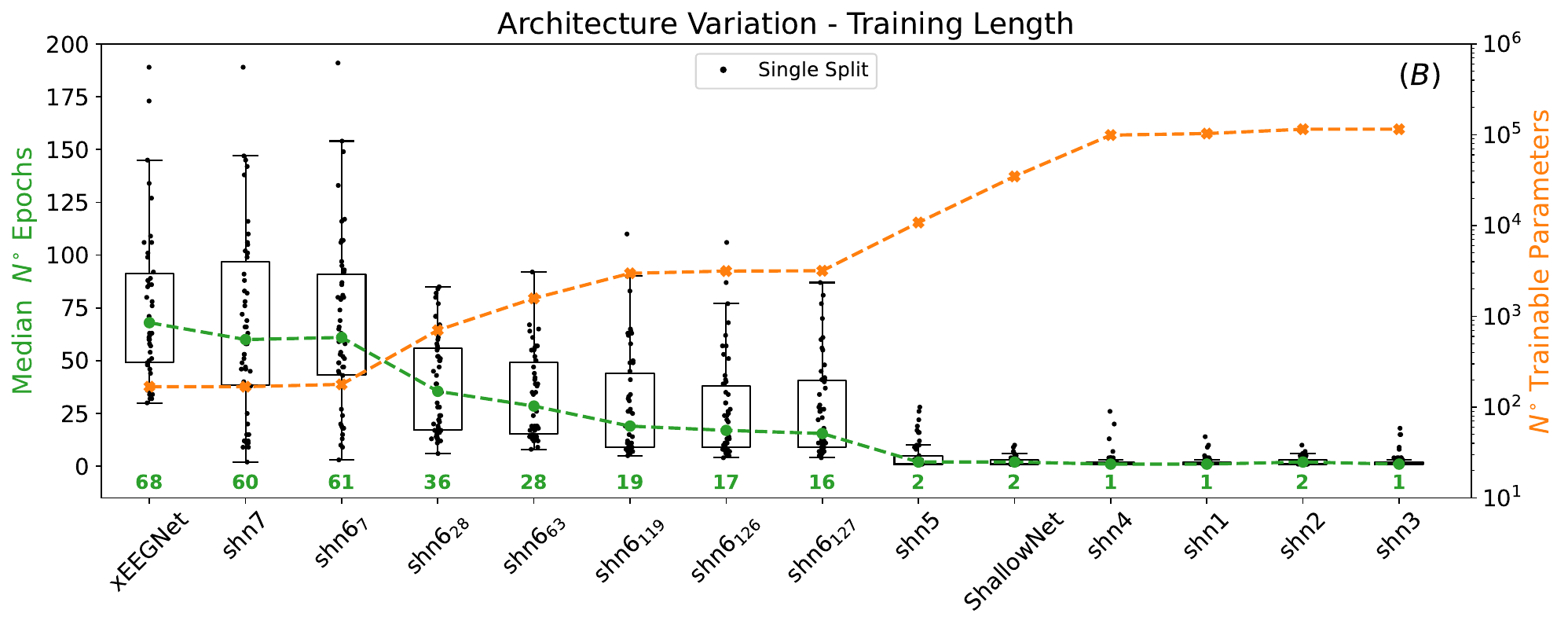}
    \caption{Architecture Variation. 
    Panel A shows how the median weighted accuracy (blue line) and the number of trainable parameters (orange line) vary with the various models reported in \autoref{tab:weights}.
    The black dots are the weighted accuracies of all the 50 train-validation-test splits inside the N-LNSO procedure.
    In red the random guess baseline for a 3-class classification problem.
    Panel B shows how the median number of epochs (green line) and the number of trainable parameters (orange line) vary considering the same models before.
    The black dots are number of epochs of all the 50 train-validation-test splits inside the N-LNSO procedure.
    The number of epochs here reported are the ones associated to the best models found and retrieved with Early Stopping.
    While the performance does not correlate with the logarithm of the number of trainable parameters, the number of epochs significantly does ($\rho(12)=-.93,~p<.001$, $Adj.~R^2=.86$).}
    \label{fig: architecture_var}
\end{figure}

\noindent
In \autoref{fig: architecture_var} the performances of the models are reported; as it can be seen in Panel A, performance does not correlate with the logarithm of the number of trainable parameters, while the number of epochs does significantly ($\rho(12)=-.93,~p<.001$, $Adj.~R^2=.86$) in Panel B.
This confirms that reducing trainable parameters leads to longer training times to achieve the same performance. 
The reported number of epochs corresponds to the best model selected using the validation set and retrieved via Early Stopping.

As can be seen from \autoref{fig: architecture_var} Panel A, ShallowNet performs better than xEEGNet in terms of median weighted accuracy (+1.5\%); at the same time, if mean weighted accuracy was used as the metric under consideration, things would have reverted, with xEEGNet performing slightly better than ShallowNet (+0.3\%).
These differences between the median and mean performances are due to the high variability of the N-LNSO performances of ShallowNet.
The Quartile Coefficient of Variation (\cite{QCV}, page 17) is in fact 32\% smaller in xEEGNet.

%THE CONCEPT OF EFFICIENCY
xEEGNet improves efficiency by reducing the number of parameters without sacrificing performance. 
Network efficiency can be measured using NetScore \cite{wong}, a metric that combines accuracy, parameter count, and multiply–accumulate operations (MACs), which reflect computational cost. 
xEEGNet achieves a NetScore of $\Omega=35.2$, indicating better efficiency compared to ShallowNet $(\Omega=12.5)$. 
Further details are provided in the supplementary material section 5.

\subsection{Features Inspection}
The next paragraphs are dedicated to the inspection of the features learned by xEEGNet.
The best performing split obtained is 9-2 (outer folder - inner folder) with a weighted accuracy of 74.5\%, while the worst performing split is 6-5 with 34.4\%.
In these two splits, ShallowNet performs, respectively, 62.0\% and 43.8\%.

\subsubsection{First Layer - EEG band specific filters}
\begin{figure}[h!]
    \centering
    \includegraphics[width=0.85\linewidth]{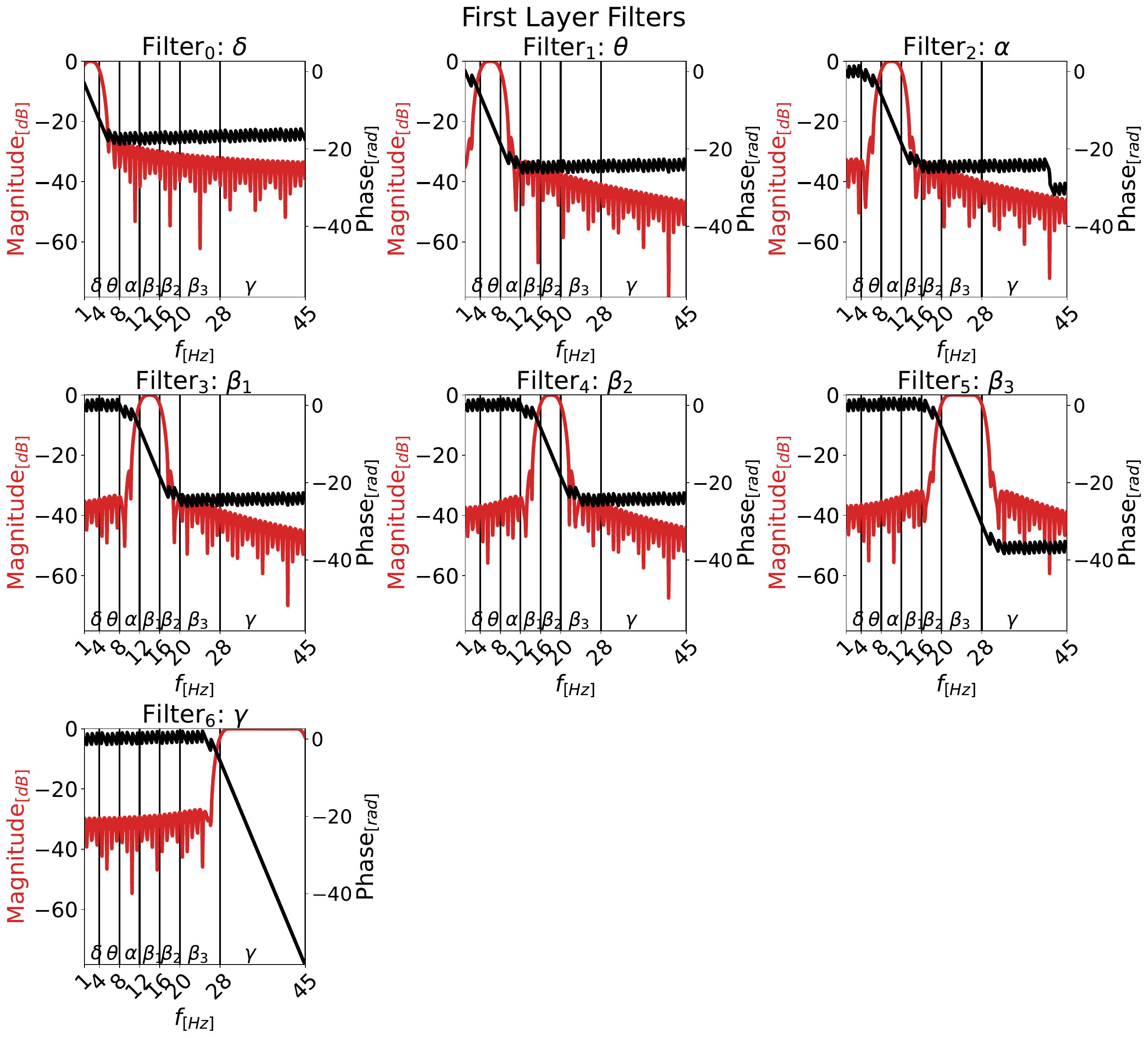}
    \caption{Filters in the first layer.
    The frequency response of the seven temporal filters in the first layer can be inspected using Bode's plots. 
    In red there is the magnitude in decibel (dB), while in black the phase measured in radians (rad); on the horizontal axis there is the frequency range divided into the seven EEG bands considered.
    Frequencies are visualized in the [1,45]Hz range.}
    \label{fig: BP}
\end{figure}

Adopting the idea of using frozen, pre-initialized and pre-defined filters, the kernels are then equal for both the best and worst split; the frequency response of these filters can be seen in \autoref{fig: BP}.

\subsubsection{Second layer - EEG band specific scalp topologies}
\begin{figure}[h!]
    \centering
    \includegraphics[width=1\linewidth]{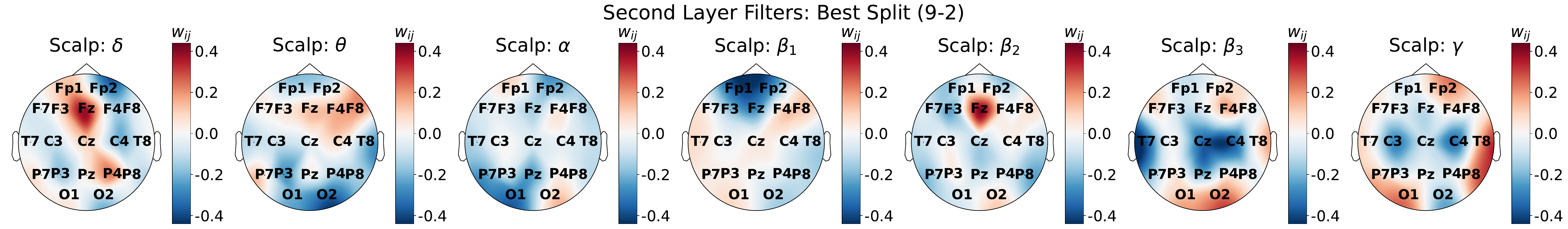}
    \vspace{0mm}
    \par
    \includegraphics[width=1\linewidth]{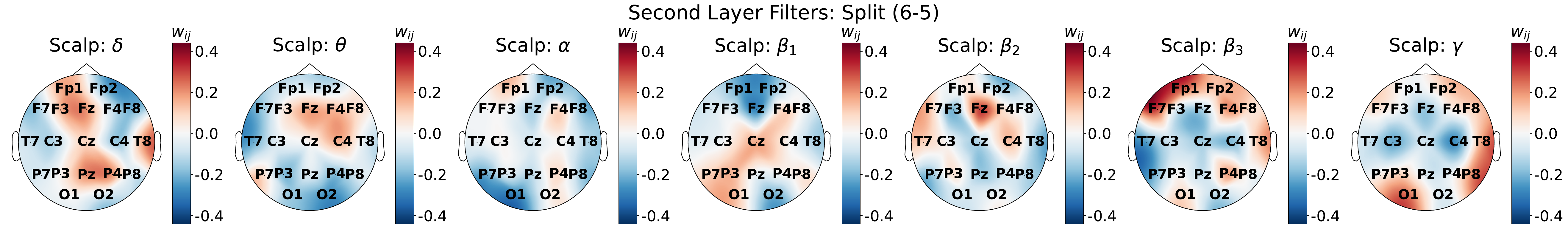}
    \caption{Scalp topographies learned in the second layer.
    On top of the figure, the set of scalp topographies learned in the best slip, while on the bottom part those learned with the worst split.
    The color-bars are equal for all the plots, indicating the value of the weights; in red there are the positive weights while in blue the negative ones.
    The scalp topographies report the weights values interpolated across the scalp surface.}
    \label{fig: scalp2}
\end{figure}

By receiving the output of the interpretable EEG band-specific filters, the second layers learn interpretable EEG-band specific scalp topologies.
The weights of the spatial kernels associated with the EEG bands for the best and worst split are shown in \autoref{fig: scalp2}.

\noindent
The similarity between the topographies for the best and worst splits can be compared using pairwise correlations.
Holm's correction \cite{holm} was used to account for multiple-comparisons in order to control the Family Wise Error Rate.
It turns out that the weights for the best and worst splits correlate significantly band-wise ($\rho(17)>.77,~p<.05$ Holm corrected, $Adj.~R^2>.57)$ except for $\beta_3$ ($\rho(17)=.59,~p=.16$).
Indeed, the $\beta_3$ topography for the best split has large weights (in absolute terms) predominantly on Cz-C4 (central-right area) and on O1-O2 (occipital area) while the worst split is mainly on F7-Fp1 (inferior-frontal and fronto-polar areas).
At the same time, two topographies of different bands and different splits do not correlate. %($p>0.05$).
Finally, within the same split, scalp topographies between different EEG bands do not correlate, making each scalp topography specific to the band involved.

\subsubsection{Global Average Pooling - Signal's power spectrum}
\begin{figure}[h!]
    \centering
    \includegraphics[width=1\linewidth]{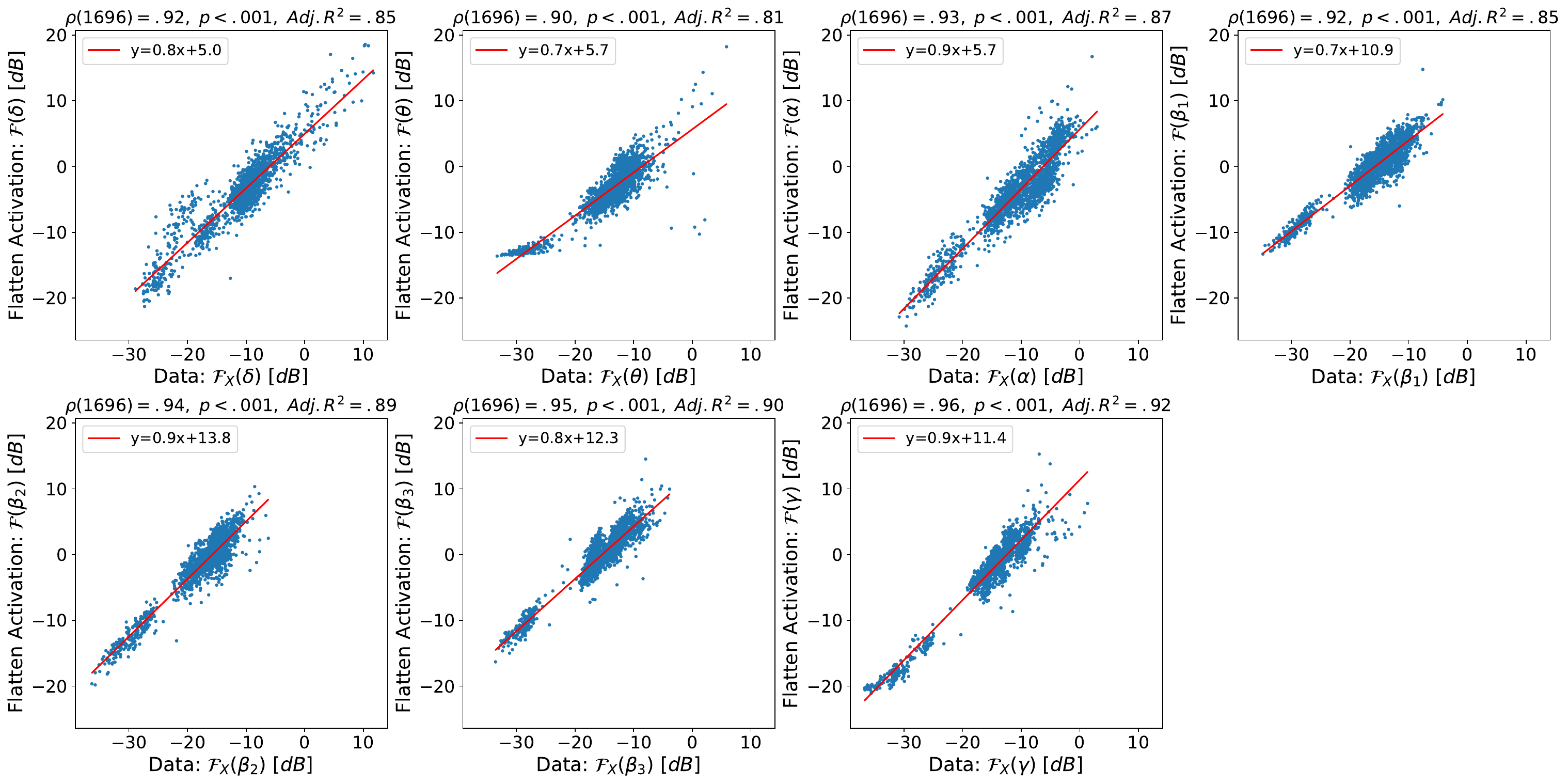}
    \caption{Correlation between activations at the beginning of the dense layer and average (over channels) band powers for the test set.
    The figure consists of seven panels, one for each EEG band.
    The x-axis shows the average band power $\mathcal{F}_X$ defined in \autoref{equ: Fx} calculated directly on the EEG windows.
    The y-axis shows the flatten activations $\mathcal{F}$ of xEEGNet.
    Values of both axis are expressed in decibel (dB).
    Blue dots indicate the input windows of the test set, while the red line is the linear fitted model.
    Axis limits are equal for all the seven sub-plots.}
    \label{fig: PSD_corr}
\end{figure}

Using global average pooling confirms that the architectures can access the signal power spectrum content, as mentioned in \autoref{softmax-activ}. 
For visualization, results are shown only for the worst split (6-5) in \autoref{fig: PSD_corr}.
The activations at the beginning of the dense layer $\mathcal{F}^n$ and the average band powers from the data $\mathcal{F}_X^n$ show a strong correlation across all seven EEG bands ($\rho(1696)>.90,~p<.001$, Holm corrected, $Adj.~R^2>.81)$.
Additionally, the slope coefficients are close to one, indicating a small distortion between the calculated $\mathcal{F}$ and extracted counterpart.
Notably, the flatten activations for all seven bands span approximately the same range of values. 
Given that lower EEG bands have greater power than higher ones, achieving this uniform interval requires adding a constant that is smaller for slower EEG bands and larger for faster ones, precisely the function of the intercept in these linear models.
This effect of having different EEG bands with approximately the same range of values is due to the batch normalization layer (see \autoref{sec: batch}).

\subsubsection{Dense Layer - Spectral bands for disease classification}
\label{sec: denselayer}
\begin{figure}[h!]
    \centering
    \includegraphics[width=0.49\linewidth]{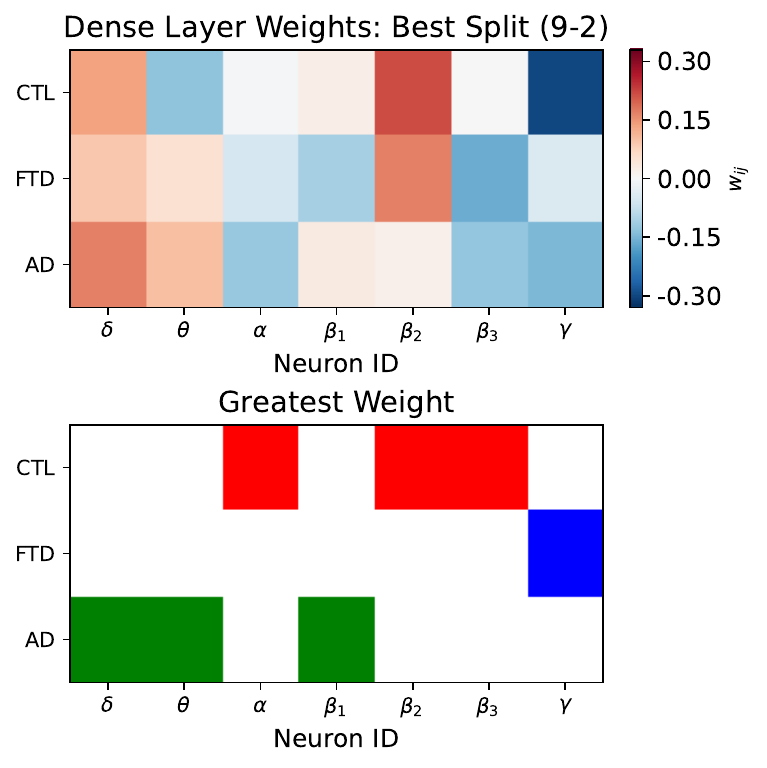}     \includegraphics[width=0.49\linewidth]{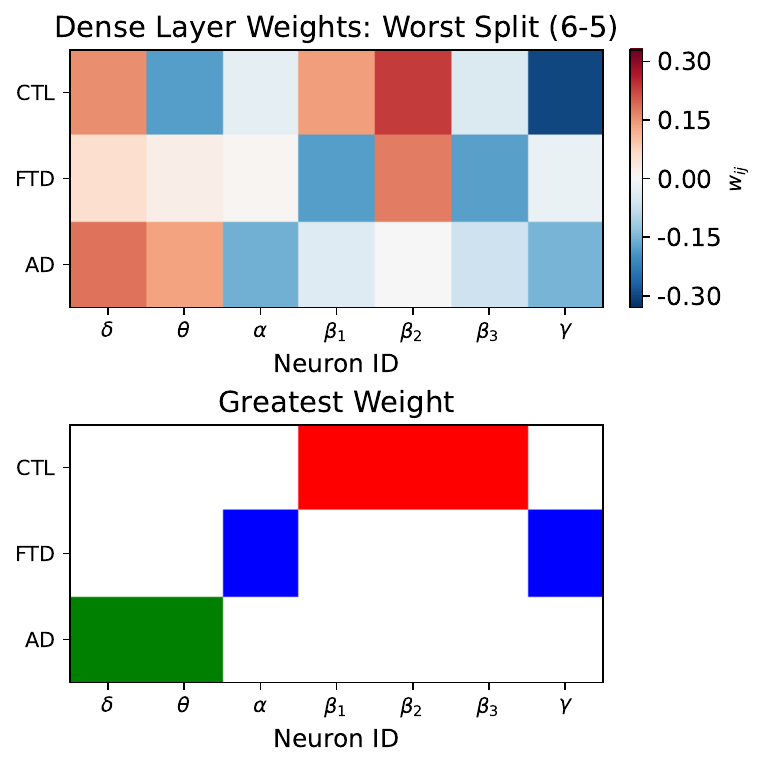}
    \caption{Weights learned in the dense layer.
    The upper part of the figure shows the weights learned in the dense layer, for the best split (left) and the worst split (right).
    The colour bars are equal for both plots and indicate the weights values, in red the positive and in blue the negative ones.
    The y-axis indicates the class label, where `CTL' indicates control, `FTD' frontotemporal dementia and `AD' Alzheimer's disease.
    In the x-axis is reported the EEG bands investigated.
    In the lower part of the figure it is shown, the largest positive weight for each class and EEG band, for the best split (left) and the worst split (right).}
    \label{fig: dense}
\end{figure}

Finally, the dense layer weights can be inspected as well and they are shown in \autoref{fig: dense}.
The figure gives valuable insight into how EEG bands are used to predict the pathology; some of them are always used to predict a particular class, regardless of the split.
While \autoref{fig: dense} only reports the best and worst splits, running all 50 splits in the N-LNSO procedure allows one to show that:
\begin{itemize} 
    \item[--] $\delta$ band (0,4] Hz: predominantly used to predict the Alzheimer's class, except in one split where it is used for control.
    \item[--] $\theta$ band (4,8] Hz: used consistently to predict the Alzheimer's class.
    \item[--] $\alpha$ band (8,12] Hz: used to predict frontotemporal dementia in 10 splits and control in 40 splits.
    \item[--] $\beta_1$ band (12,16] Hz: primarily used to predict the control class, except in 3 splits where it is used for Alzheimer's.
    \item[--] $\beta_2$ band (16,20] Hz: always used to predict the control class.
    \item[--] $\beta_3$ band (20,28] Hz: used to predict frontotemporal dementia in 13 splits and control in 37 splits.
    \item[--] $\gamma$ band (28,45] Hz: always used to predict the frontotemporal dementia class.
\end{itemize}
This aligns to known literature \cite{Dauwels, Rossini} and with the results reported in Zanola \etal~\cite{Zanola_2024} which shows how delta and theta bands are significantly higher in Alzheimer's patients, for the same dataset considered in this work.
At the same time, alpha and beta bands are significantly higher and associated to the healthy, control condition.
Finally the gamma band, is significantly higher in patients from the frontotemporal dementia's class.

\subsection{Evaluating xEEGNet for Dementia Classification}
\subsubsection{Overfitting}
% The theoretical phenomenon of overfitting was introduced in \autoref{sec: overfitting}, which illustrated how it can be formulated into two possible frameworks.
% One way to study and quantify overfitting is to look at the number of epochs of training.
% \autoref{fig: architecture_var} Panel B, shows that big models, usually employ very few epochs to achieve the best configuration.
% When models decrease in size, the number of epochs required increase; notice that in every model under test, except the proposed one xEEGNet, there is at least one split inside the N-LNSO strategy, that requires one epoch of training.
% The second way to study overfitting is through the train and validation loss curves, their trends and the respective correlation.
% It can be clearly seen on the central panels of \autoref{fig: curve_loss}, where while ShallowNet and `shn5' (black and brown lines) have an increasing validation loss, a clear sign of overfitting, the others models have not.
% At the same time, right panels shows clearly that `shn$_{127}$', shn$_{28}$ and xEEGNet (red, orange and green lines) show negative correlations between training and validation loss; notice that also here, our proposed model outperforms the others, with the most skewed distribution.
% It is worth to underline that the transition between a negative correlations regime (brown model) to a positive one (red model), is due to the introduction of the global average pooling, and the consequent reduction of parameters involved in the dense layer.

The theoretical concept of overfitting, introduced in \autoref{sec: overfitting}, was formulated into two possible frameworks.
One way to study overfitting is by analyzing the number of training epochs. \autoref{fig: architecture_var}, Panel B, shows that larger models require fewer epochs to reach optimal performance. 
As model size decreases, the required epochs increase. 
Notably, every tested model, except the proposed xEEGNet, has at least one N-LNSO split requiring just one epoch to reach the highest validation accuracy.

\begin{figure}[h!]
    \centering
    \includegraphics[width=1\linewidth]{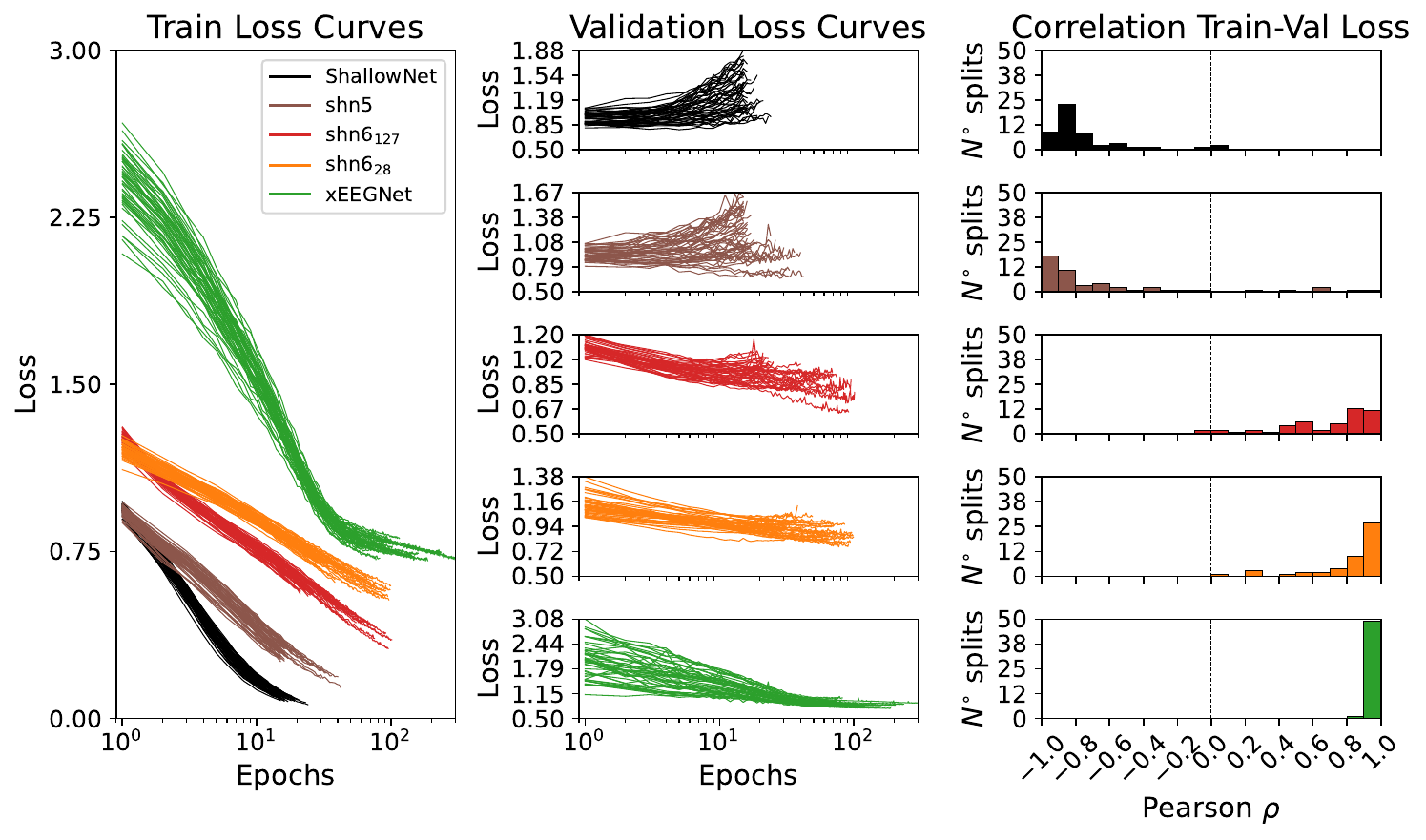}
    \caption{Training and validation loss curves with respective correlations.
    Five models from \autoref{tab:weights} have been selected respectively: ShallowNet (black line), `shn5' (brown line), `shn$6_{127}$' (red line), `shn$6_{28}$' (orange line) and xEEGNet (green line) (see \autoref{tab:weights}).
    For each model, in the left side of the figure it is reported the training loss curves for all the 50 train-validation-test splits given by the N-LNSO procedure.
    At the center of the figure instead, there are the validation loss curves for the five models, one for each sub-plot.
    On the right side, the distribution of the Pearson correlation coefficients between the training and validation loss for the 50 splits of the N-LNSO strategy.
    It can be seen that ShallowNet and `shn5' (black and brown lines) display overfitting, with an increasing validation loss, which is negatively correlated with the training loss.}
    \label{fig: curve_loss}
\end{figure}

A second approach is to examine the train and validation loss curves. While in \autoref{fig: curve_loss} left panel shows that train loss monotonically decrease for all models, the central panels shows that ShallowNet and `shn5' (black and brown lines) exhibit increasing validation loss, indicating overfitting, whereas other models do not. 
The right panels further highlight that `shn$6_{127}$', `shn$6_{28}$', and xEEGNet (red, orange, and green lines) have negative correlations between training and validation loss. 
Among them, xEEGNet shows the most skewed distribution, outperforming the rest.
This shift from negative (brown model) to positive correlation (red model) results from introducing global average pooling, which reduces the number of dense layer parameters.

\subsubsection{Neasted-Leave-N-Subjects Out Performances}
\label{sec: feasibility}
\begin{figure}[h!]
    \centering
    \includegraphics[width=0.55\linewidth]{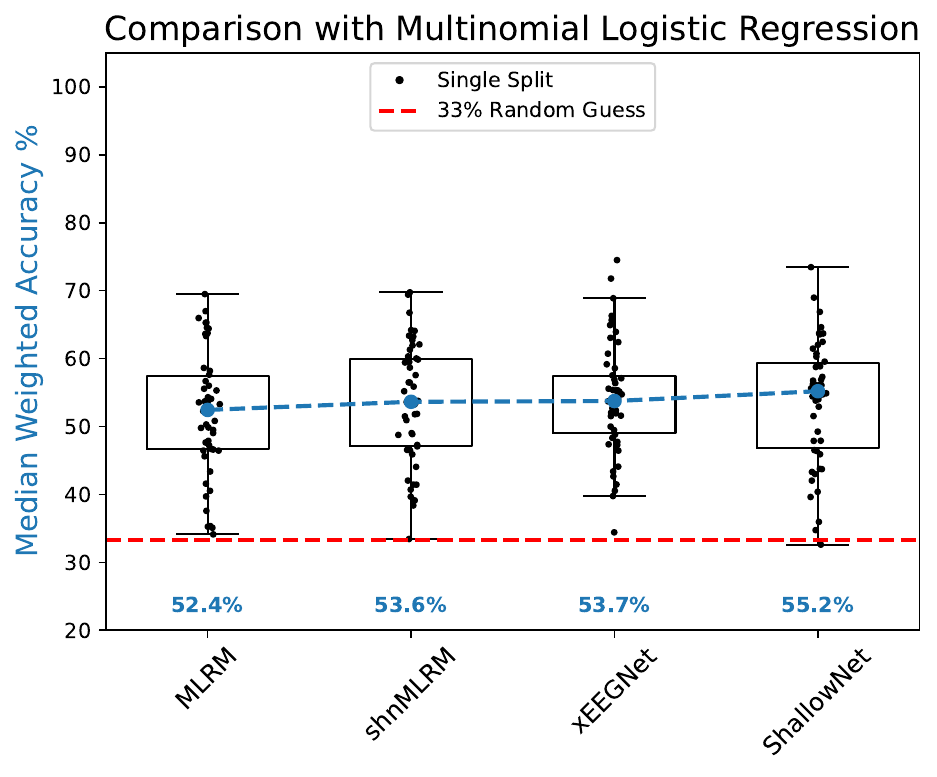}
    \caption{Multinomial Logistical Regression results.
    This figure shows the median weighted accuracy (blu line) for the four different models investigated.
    `MLRM' stays for multinomial logistic regression model, which predict the class based on the average band power $\mathcal{F}_X$ (see \autoref{equ: Fx}).
    `shnMLRM' is an hybrid solution, which uses the convolutional encoder of xEEGNet and substitute the dense layer with a MLRM.
    For xEEGNet and ShallowNet the boxplots reported are the same of \autoref{fig: architecture_var}, Panel A.
    In red the random guess baseline for a 3-class classification problem.}
    \label{fig: MLR}
\end{figure}

The four models listed in \autoref{sec: N-LNSO performances} (ShallowNet, xEEGNet, MLRM, `shnMLRM') have been trained with a N-LNSO strategy and the results are reported in \autoref{fig: MLR}.
First, ShallowNet, a deep neural network with nearly \num{35000} parameters, outperforms the simple MLR model, achieving a +2.8\% improvement in median weighted accuracy.
Additionally, ‘shnMLRM’ shows a +1.2\% increase over MLRM, suggesting that the primary factor behind xEEGNet surpassing the statistical model (+1.3\%) is not the extraction of the EEG bands, but the use of the dense layer.
If the mean weighted accuracy was used instead, the gap between MLRM (52.0\%) and ShallowNet (53.6\%) would shrink to just +1.6\%.
Although MLRM's performance is modest, it is comparable to ShallowNet, whose reliance on similar low-level features similarly constrains performance, indicating that the three-class pathology classification task cannot be solved solely using the average power of the seven EEG bands.
Thus, MLRM may serve as a lower bound for this three-class dementia classification task.
These results encourage comparing deep learning models with statistical counterparts whenever possible.

% Again, if the mean weighted accuracy would be used instead as metric under consideration, the difference between MLRM (52.0\%) and ShallowNet (53.6\%) would be reduced to only +1.6\%.
% Since the performance of the MLRM is still not great, but undoubtedly comparable with ShallowNet, this could suggest that the task of pathology classification cannot be solved exclusively using the average power of the seven EEG bands.
% The MLRM performance thus could be seen as the lower bound for this 3-class classification problem of dementia diagnosis.
% We encourage the practice, whenever possible, to compare DL model performances with statical model counterparts.

\subsubsection{Impact of Data Partitioning on Performance Variability}
\label{sec: N-LNSO Variability res}
The weighted accuracy significantly correlates $(p<.05,~\text{Holm corrected})$, with only 5 of the 36 separability defined by \autoref{equ: separability} in \autoref{sec: N-LNSO Variability}; details are reported in \autoref{tab: corr}.
\begin{table}[h!]
    \centering
    \begin{tabular}{c|c|c}
         Separability & $\rho(48)$ & $p_{value}$  \\
         \hline
         \hline
         $\text{FTD}_{tr}~\text{-}~\text{CTL}_{ts}$   & .49 & .01 \\
         \hline
         $\text{CTL}_{tr}~\text{-}~\text{AD}_{ts}$   & .56 & $<$.001 \\
        \hline
         $\text{CTL}_v~\text{-}~\text{AD}_{ts}$   & .51 & .005 \\
          \hline
         $\text{AD}_{ts}~\text{-}~\text{CTL}_{ts}$   & .85 & $<$.001 \\  
         \hline
         $\text{FTD}_{ts}~\text{-}~\text{CTL}_{ts}$   & .47 & .02 \\        
    \end{tabular}
    \caption{Significant correlations between weighted accuracy and the 36 separability.
    Parsons' correlation coefficients $\rho$ are reported together with the respective p-values (Holm corrected).
    $tr,~v,~ ts$ stays for training, validation and test set.
    CTL, FTD and AD stays for control, frontotemporal dementia and Alzheimer.}
    \label{tab: corr}
\end{table}
These variables are then passed to the Forward-Backward Feature Selection algorithm, which gives an optimal regression model composed of only one independent feature, that is, $\text{AD}_{ts} - \text{CTL}_{ts}$. %; the VIF of all the selected variables is less than 5, showing low multi-collinearity.
The regression through ordinary least squares gives a linear model highly descriptive of the variability of the performances $(\text{Adj.}~R^2=.715,~F(1,48)=123.9,~p<.001,~AIC=-168)$.
The fitted linear model is described in \autoref{tab: coef}; scatter-plots between the selected variables and the weighted accuracy is reported in the supplementary material section 6.
\begin{table}[h!]
    \centering
    \begin{tabular}{c|c|c|c|c|c c}
         & coef  &  std err & $t$ & $P>|t|$ &  $[$0.025  &  0.975$]$ \\
         \hline
         \hline
        Intercept             &    0.3834  &    0.015   &  25.034   &   $<$.001   &    0.353  &     0.414 \\
        $\text{AD}_{ts}~\text{-}~\text{CTL}_{ts}$    &   0.2375  &    0.021   &  11.130   &   $<$.001   &   0.195  &    0.280 \\
    \end{tabular}
    \caption{Model coefficients found by the regression fit.
    Coef stays for coefficient; std err stays for standard error.
    $tr,~v,~ ts$ stays for training, validation and test set.
    CTL, FTD and AD stays for control, frontotemporal dementia and Alzheimer.}
    \label{tab: coef}
\end{table}

\noindent
In summary, the variability of the weighted accuracy is described by the following:
\begin{itemize}
    \item[--] The greater the separation between $\text{CTL}_{ts}$ (controls from test) and $\text{AD}_{ts}$ (Alzheimer's from test) the higher the accuracy.
\end{itemize}

%%% WHAT IS INTRESTING IS WHAT IS NOT DRIVING THE VARIABILITY %%%
The separability between the two major classes in the test set is expected to be the primary factor influencing the classification performance. 
However, the proposed methodology approaches this issue from the perspective of the topological representation learned by the network.
As shown in \autoref{tab: corr}, the separability between training and test sets, computed for each class, does not exhibit a significant correlation with performance. 
Nonetheless, the average separability between corresponding training and test sets for controls ($\text{CTL}_{tr}$ - $\text{CTL}_{ts}$), frontotemporal dementia patients ($\text{FTD}_{tr}$ - $\text{FTD}_{ts}$) and Alzheimer's patients ($\text{AD}_{tr}$ - $\text{AD}_{ts}$) remains consistently low across all 50 splits, with values of $0.25 \pm 0.14$, $0.30 \pm 0.10$ and $0.27 \pm 0.12$, respectively. 
This confirms that each training set is, on average, topologically close to its corresponding test set.
%%% WHAT IS INTRESTING IS THE DIFFERENCE WITH SHALLOWNET %%%
The same analysis was also performed for ShallowNet (see supplementary material, section 6).
Both models exhibit a strong relationship between classification accuracy and the separability between Alzheimer’s patients and controls in the test set. 
However, ShallowNet also demonstrates a dependency on separability between frontotemporal dementia samples seen during training and control samples presented in the test set.
%%% WHAT DOES IT MEAN?
This suggests that its performance depends not only on the intrinsic separability of test samples (AD vs. CTL) but also on how well frontotemporal dementia samples in training are separated from controls in the test set. 
In contrast xEEGNet, which avoids overfitting, shows no dependency on the separability of the training data.
%%% FINAL CONCLUSION ON THE METHODOLOGY
In conclusion, the variability observed in N-LNSO can be explained by the relative positioning (i.e., separability) of EEG window groups, categorized by sets, labels, and their combinations.

\section{Discussion}\label{discussion}
% DISCUSSION
\label{sec-discussion}

%GENERAL INTRODUCTION
The foundation of this work rests on the definition of interpretability proposed by Miller \etal~in \cite{MILLER20191}, where interpretability is described as the degree to which a human can understand the reasoning behind a model’s decision. 
This paper was developed with the aim of achieving such interpretability. 
In particular, \autoref{sec-networkdescription} provides insights into the network structure, setting the stage for theoretically motivated modifications in \autoref{sec-networkmodification}. 
These modifications are not intended to improve performance but rather to maintain it at a comparable level, while increasing interpretability.

%FIRST POINT: IMPORTANCE OF SMALL MODELS
One of the most significant observations of this study is that the size of an architecture does not necessarily correlate with its performance.
If larger models do not yield substantial improvements over smaller ones, the latter should be prioritized. 
Small models produce simpler hypotheses, which aligns with Occam's Razor, suggesting simpler models are preferable. 
In addition, larger models pose a serious risk of overfitting.
Early Stopping identified optimal models in the first or second epoch for larger architectures, such as ShallowNet, `shn1', `shn2', `shn3', `shn4', and `shn5', as shown in \autoref{fig: architecture_var} (Panel B).
This pattern indicates that these architectures may be memorizing the training set instead of learning the true function \( f:\mathcal{D} \rightarrow \mathcal{L} \). 
Supporting this notion is the fact that the dataset consists of only \num{17604} samples, which is fewer than the number of parameters in architectures like ShallowNet (\num{34803} parameters). 
According to the heuristic of the ``10-times rule", a deep neural network should ideally be trained with a number of samples ten times the size of the model's parameters.
The model `shn$6_{28}$' (orange line in \autoref{fig: curve_loss}), with only \num{703} parameters, is the first among the models to exhibit a skewed distribution of Pearson's correlations toward one, indicating the absence of overfitting.

%SECOND POINT: WE KNOW WHAT IS DOING
The modifications applied to ShallowNet resulted in the proposed model, xEEGNet, which effectively mitigates overfitting and yields an interpretable analytical function $f$. 
Although ShallowNet was already well understood and documented by its authors \cite{shallow}, this understanding did not directly translate into interpretability for the end user. 
In contrast, xEEGNet offers a clear rationale behind its EEG window pathology predictions. 
For example, if an EEG window is classified as Alzheimer's, \autoref{equ: class_pred} reveals that the model's decision is based on the fact that the weighted sum of the power bands is higher for that class. 
The weights in the dense layer in \autoref{fig: dense} are medically significant as they align with medical literature, indicating that the prediction is based on spectral content consistent with the pathology’s profile. 
Directly supported by \autoref{fig: PSD_corr} and indirectly by \autoref{fig: MLR}, xEEGNet encoder accurately extracts the spectral content from the time windows. 
This accuracy is enabled by the depthwise convolutional layer, which preserves individual EEG bands, and by the averaging of signals after filtering.

%THIRD POINT: PSD IS NOT ENOUGH
While xEEGNet performs comparably to ShallowNet, their overall performance remains limited. 
Furthermore, a simple Multinomial Logistic Regression (MLR) model underperforms only by -2.6\%, raising questions about the need for deep neural networks for this task. 
Moreover, the best practice of comparing deep learning models with traditional statistical methods is suggested and highly recommended to illustrate any substantial performance gain.
Furthermore, the results in \autoref{sec: N-LNSO Variability res}, propose a methodology to investigate the reasons behind the model's performance variability; while the characteristics found to be relevant for the performance variability remain specific for the deep learning model studied and dataset considered, the methodology proposed is independent of them.
This provides another tool to understand the reasons behind the model performance.

% 3.5 POINT: LIMITATIONS
A potential criticism of xEEGNet is its defining feature: the extremely low number of parameters (only 168 for the dataset and task considered). 
While this minimalist architecture offers improvements over the original ShallowNet, it may not be seen as a truly ``deep" network. 
Its limited complexity and representational capacity raise concerns about its applicability beyond neurodegenerative pathology classification, where performance may remain modest.
Nevertheless, xEEGNet could serve as a sub-module within a larger architecture, contributing its strength in extracting the signal’s power spectrum. 
This could be complemented by modules focused on other spatio-temporal features, such as channel connectivity. 
These features could be fused using attention mechanisms, as in transformer-based models \cite{eegconformer}.
Alternatively, xEEGNet could be trained via knowledge distillation \cite{distillation}, where a larger, more complex teacher network guides its learning. 
This would combine the interpretability and efficiency of the student model with the expressive power of the teacher.

%FOURTH POINT: EEG as TOOL FOR DIAGNOSIS
Although algorithms that classify pathologies based on short EEG windows are promising, it is essential to recognize why current performance may fall short.
As suggested in the medical literature, Alzheimer’s diagnosis is rarely based solely on the spectral power of EEG bands from resting-state recordings.
In 2010, Dauwels \etal~\cite{insuffEEG} noted the difficulty of evaluating the utility of EEG for Alzheimer's diagnosis, while more recently, Babiloni \etal~\cite{babiloni}, 2016, speculated that the EEG power density could support, rather than replace, standard assessments for conditions such as mild cognitive impairment (MCI) and Alzheimer’s disease (AD). 
This shift in time acknowledges the supportive role of EEG in diagnosis.
This work further emphasizes the importance of multimodal learning for pathology classification, particularly in neurodegenerative diseases like Alzheimer’s and frontotemporal dementia. 

%FIFTH POINT: NECESSITY TO MULTI-MODAL LEARNING and FUTURE WORK
Given that spectral power alone is insufficient, multimodal approaches can be pursued by integrating cognitive data, such as the MDS-UPDRS scale \cite{UPDRS}, MMSE \cite{MMSE}, and MOCA \cite{MOCA} scores, or structural MRI biomarkers, along with EEG-derived features as done in \cite{multimodalEEG}.
Other modalities, including MRI, PET, and genetic data, have also been effectively combined with EEG data \cite{multimodalEEG1}. 
Such multimodal frameworks could offer in the future a more comprehensive view of the clinical states of patients, which can lead to more accurate diagnoses.

\section{Conclusion}\label{conclusions}
% CONCLUSIONS
This paper presents xEEGNet, highlighting the importance of interpretability in deep learning models for dementia classification. 
Starting from ShallowNet, we transform a ``black box'' model into a ``white box'' one by progressively modifying the layers to increase network interpretability. 
xEEGNet maintains comparable performance while reducing variability across data splits and offering clear medical insight into how pathology is predicted from EEG windows.
Hence, this smaller model enables simpler and more interpretable hypotheses without sacrificing performance; it resists overfitting and can be implemented with limited computational resources. 
Comparison with multinomial logistic regression further underscores the inherent limitations of predicting pathology using only spectral features, emphasizing the need to integrate EEG with other modalities to improve classification accuracy.
 
\section{Acknowledgment}
\begin{itemize}
    \item Funding: This work was supported by the STARS@UNIPD funding program of the University of Padova, Italy, through the project: MEDMAX. 
    This project has received funding from the European Union’s Horizon Europe research and innovation programme under grant agreement no 101137074 - HEREDITARY.
    \item Conflict of interest: The authors have no competing interests to declare that are relevant to the content of this article.
    \item Code availability: The code is available at \href{https://github.com/MedMaxLab/shallownetXAI}{MedMaxLab/shallownetXAI}.
    \item Authors' contributions: AZ - Conceptualization, Methodology, Model Training, Statistical Analysis, Writing - Original Draft. 
    LFT - Methodology, Writing - Original Draft.
    FDP - Methodology.
    MB - Methodology, Statistical Analysis.
    MA - Project Administration, Supervision. 
    All authors reviewed and edited previous versions of the manuscript.
    \item Data availability: All data supporting the findings of this study are openly available within the OpenNeuro platform.
\end{itemize}

\section{Bibliography}
\bibliography{bibliography}

\providecommand{\newblock}{}
\begin{thebibliography}{10}
\expandafter\ifx\csname url\endcsname\relax
  \def\url#1{{\tt #1}}\fi
\expandafter\ifx\csname urlprefix\endcsname\relax\def\urlprefix{URL }\fi
\providecommand{\eprint}[2][]{\url{#2}}

\bibitem{SSL_EEG}
Rafiei M~H, Gauthier L~V, Adeli H and Takabi D 2022 Self-supervised learning for electroencephalography {\em IEEE Transactions on Neural Networks and Learning Systems\/} \href{https://doi.org/10.1109/TNNLS.2022.3190448}{{\bf 35} 1457--1471}

\bibitem{Yannick}
Roy Y, Banville H, Albuquerque I, Gramfort A, Falk T~H and Faubert J 2019 Deep learning-based electroencephalography analysis: a systematic review {\em Journal of Neural Engineering\/} \href{https://doi.org/10.1088/1741-2552/ab260c}{{\bf 16} 051001}

\bibitem{Limitations}
Rubinger L, Gazendam A, Ekhtiari S and Bhandari M 2023 Machine learning and artificial intelligence in research and healthcare {\em Injury\/} \href{https://doi.org/https://doi.org/10.1016/j.injury.2022.01.046}{{\bf 54} S69--S73} ISSN 0020-1383

\bibitem{prob2018}
Miotto R, Wang F, Wang S, Jiang X and Dudley J~T 2017 {Deep learning for healthcare: review, opportunities and challenges} {\em Briefings in Bioinformatics\/} \href{https://doi.org/10.1093/bib/bbx044}{{\bf 19} 1236--1246} ISSN 1477-4054

\bibitem{prob2021}
Nisar D~E~M, Amin R, Shah N~U~H, Ghamdi M~A~A, Almotiri S~H and Alruily M 2021 Healthcare techniques through deep learning: Issues, challenges and opportunities {\em IEEE Access\/} \href{https://doi.org/10.1109/ACCESS.2021.3095312}{{\bf 9} 98523--98541}

\bibitem{prob2024}
Ciobanu-Caraus O, Aicher A, Kernbach J~M, Regli L, Serra C and Staartjes V~E 2024 A critical moment in machine learning in medicine: on reproducible and interpretable learning {\em Acta Neurochirurgica\/} \href{https://doi.org/10.1007/s00701-024-05892-8}{{\bf 166} 14} ISSN 0942-0940

\bibitem{TowardBestPractice}
Cui J, Yuan L, Wang Z, Li R and Jiang T 2023 Towards best practice of interpreting deep learning models for eeg-based brain computer interfaces {\em Frontiers in Computational Neuroscience\/} \href{https://doi.org/10.3389/fncom.2023.1232925}{{\bf 17}} ISSN 1662-5188

\bibitem{THELANCETRESPIRATORYMEDICINE2018801}
{The Lancet Respiratory Medicine} 2018 Opening the black box of machine learning {\em The Lancet Respiratory Medicine\/} \href{https://doi.org/https://doi.org/10.1016/S2213-2600(18)30425-9}{{\bf 6} 801} ISSN 2213-2600

\bibitem{Jacovi}
Jacovi A, Marasovi\'{c} A, Miller T and Goldberg Y 2021 Formalizing trust in artificial intelligence: Prerequisites, causes and goals of human trust in ai {\em Proceedings of the 2021 ACM Conference on Fairness, Accountability, and Transparency\/} FAccT '21 (New York, NY, USA: Association for Computing Machinery) p 624–635 ISBN 9781450383097

\bibitem{MILLER20191}
Miller T 2019 Explanation in artificial intelligence: Insights from the social sciences {\em Artificial Intelligence\/} \href{https://doi.org/https://doi.org/10.1016/j.artint.2018.07.007}{{\bf 267} 1--38} ISSN 0004-3702

\bibitem{shallow}
Schirrmeister R~T, Springenberg J~T, Fiederer L~D~J, Glasstetter M, Eggensperger K, Tangermann M, Hutter F, Burgard W and Ball T 2017 Deep learning with convolutional neural networks for eeg decoding and visualization {\em Human Brain Mapping\/} \href{https://doi.org/https://doi.org/10.1002/hbm.23730}{{\bf 38} 5391--5420}

\bibitem{eegnet}
Lawhern V~J, Solon A~J, Waytowich N~R, Gordon S~M, Hung C~P and Lance B~J 2018 {EEGNet}: a compact convolutional neural network for {EEG}-based brain--computer interfaces {\em Journal of neural engineering\/} \href{https://doi.org/10.1088/1741-2552/aace8c}{{\bf 15} 056013}

\bibitem{eegprepro}
Del~Pup F, Zanola A, Fabrice~Tshimanga L, Bertoldo A and Atzori M 2025 The more, the better? evaluating the role of eeg preprocessing for deep learning applications {\em IEEE Transactions on Neural Systems and Rehabilitation Engineering\/} \href{https://doi.org/10.1109/TNSRE.2025.3547616}{{\bf 33} 1061--1070}

\bibitem{BORRA202055}
Borra D, Fantozzi S and Magosso E 2020 Interpretable and lightweight convolutional neural network for eeg decoding: Application to movement execution and imagination {\em Neural Networks\/} \href{https://doi.org/https://doi.org/10.1016/j.neunet.2020.05.032}{{\bf 129} 55--74} ISSN 0893-6080

\bibitem{Xception}
Chollet F 2017 Xception: Deep learning with depthwise separable convolutions {\em 2017 IEEE Conference on Computer Vision and Pattern Recognition (CVPR)\/} pp 1800--1807

\bibitem{BCIdataset}
Tangermann M, Müller K~R, Aertsen A, Birbaumer N, Braun C, Brunner C, Leeb R, Mehring C, Miller K~J, Mueller-Putz G, Nolte G, Pfurtscheller G, Preissl H, Schalk G, Schlögl A, Vidaurre C, Waldert S and Blankertz B 2012 Review of the bci competition iv {\em Frontiers in Neuroscience\/} \href{https://doi.org/10.3389/fnins.2012.00055}{{\bf 6}} ISSN 1662-453X

\bibitem{ds004504}
Miltiadous A, Tzimourta K~D, Afrantou T, Ioannidis P, Grigoriadis N, Tsalikakis D~G, Angelidis P, Tsipouras M~G, Glavas E, Giannakeas N and Tzallas A~T 2023 A dataset of {EEG} recordings from: {Alzheimer's disease}, frontotemporal dementia and healthy subjects \url{https://doi.org/10.18112/openneuro.ds004504.v1.0.6}

\bibitem{OpenNeuro}
Markiewicz C~J, Gorgolewski K~J, Feingold F, Blair R, Halchenko Y~O, Miller E, Hardcastle N, Wexler J, Esteban O, Goncavles M, Jwa A and Poldrack R 2021 The {OpenNeuro} resource for sharing of neuroscience data {\em eLife\/} \href{https://doi.org/10.7554/eLife.71774}{{\bf 10} e71774}

\bibitem{casesAD}
Miltiadous A, Tzimourta K~D, Giannakeas N, Tsipouras M~G, Afrantou T, Ioannidis P and Tzallas A~T 2021 Alzheimer’s disease and frontotemporal dementia: A robust classification method of eeg signals and a comparison of validation methods {\em Diagnostics\/} \href{https://doi.org/10.3390/diagnostics11081437}{{\bf 11}} ISSN 2075-4418

\bibitem{DICE-net}
Miltiadous A, Gionanidis E, Tzimourta K~D, Giannakeas N and Tzallas A~T 2023 Dice-net: A novel convolution-transformer architecture for alzheimer detection in eeg signals {\em IEEE Access\/} \href{https://doi.org/10.1109/ACCESS.2023.3294618}{{\bf 11} 71840--71858}

\bibitem{LOSO}
Kunjan S, Grummett T~S, Pope K~J, Powers D~M~W, Fitzgibbon S~P, Bastiampillai T, Battersby M and Lewis T~W 2021 The necessity of leave one subject out (loso) cross validation for eeg disease diagnosis {\em Brain Informatics: 14th International Conference, BI 2021, Virtual Event, September 17–19, 2021, Proceedings\/} (Berlin, Heidelberg: Springer-Verlag) p 558–567 ISBN 978-3-030-86992-2

\bibitem{losoF}
Del~Pup F, Zanola A, Tshimanga L~F, Bertoldo A, Finos L and Atzori M 2025 The role of data partitioning on the performance of eeg-based deep learning models in supervised cross-subject analysis: a preliminary study {\em Preprint submitted to Elsevier\/}

\bibitem{Dauwels}
Dauwels J, Vialatte F and Cichocki A 2010 Diagnosis of {Alzheimer's disease from EEG} signals: Where are we standing? {\em Current Alzheimer Research\/} \href{https://doi.org/10.2174/156720510792231720}{{\bf 7} 487--505}

\bibitem{ReviewInterp}
Salahuddin Z, Woodruff H~C, Chatterjee A and Lambin P 2022 Transparency of deep neural networks for medical image analysis: A review of interpretability methods {\em Computers in Biology and Medicine\/} \href{https://doi.org/https://doi.org/10.1016/j.compbiomed.2021.105111}{{\bf 140} 105111} ISSN 0010-4825

\bibitem{Jasper}
Jasper H~H 1958 Ten-twenty electrode system of the international federation {\em Electroencephalogr Clin Neurophysiol\/} {\bf 10} 371--375

\bibitem{Zanola_2024}
Zanola A, Pup F~D, Porcaro C and Atzori M 2024 Bidsalign: a library for automatic merging and preprocessing of multiple eeg repositories {\em Journal of Neural Engineering\/} \href{https://doi.org/10.1088/1741-2552/ad6a8c}{{\bf 21} 046050}

\bibitem{PyTorch}
Paszke A, Gross S, Massa F, Lerer A, Bradbury J, Chanan G, Killeen T, Lin Z, Gimelshein N, Antiga L, Desmaison A, Kopf A, Yang E, DeVito Z, Raison M, Tejani A, Chilamkurthy S, Steiner B, Fang L, Bai J and Chintala S 2019 Pytorch: An imperative style, high-performance deep learning library {\em Advances in Neural Information Processing Systems\/} vol~32 ed Wallach H, Larochelle H, Beygelzimer A, d\textquotesingle Alch\'{e}-Buc F, Fox E and Garnett R (Curran Associates, Inc.) (arXiv:\href{https://doi.org/10.48550/arXiv.1912.01703}{1912.01703})

\bibitem{earlystopping}
Morgan N and Bourlard H 1989 Generalization and parameter estimation in feedforward nets: some experiments {\em Proceedings of the 2nd International Conference on Neural Information Processing Systems\/} NIPS'89 (Cambridge, MA, USA: MIT Press) p 630–637

\bibitem{selfeeg}
Del~Pup F, Zanola A, Tshimanga L~F, Mazzon P~E and Atzori M 2024 {SelfEEG}: A {Python} library for self-supervised learning in electroencephalography {\em Journal of Open Source Software\/} \href{https://doi.org/10.21105/joss.06224}{{\bf 9} 6224}

\bibitem{Wang}
Wang Y, McCane B, McNaughton N, Huang Z, Shadli h and Neo P 2019 Anxietydecoder: An eeg-based anxiety predictor using a 3-d convolutional neural network {\em 2019 International Joint Conference on Neural Networks (IJCNN)\/} pp 1--8

\bibitem{batchnorm}
Ioffe S and Szegedy C 2015 Batch normalization: accelerating deep network training by reducing internal covariate shift {\em Proceedings of the 32nd International Conference on International Conference on Machine Learning - Volume 37\/} ICML'15 (JMLR.org) p 448–456

\bibitem{dropout}
Srivastava N, Hinton G, Krizhevsky A, Sutskever I and Salakhutdinov R 2014 Dropout: A simple way to prevent neural networks from overfitting {\em Journal of Machine Learning Research\/} \href{http://jmlr.org/papers/v15/srivastava14a.html}{{\bf 15} 1929--1958}

\bibitem{parseval}
Stoica P and Moses R~L 2005 {\em Spectral analysis of signals\/} (Upper Saddle River, N.J.: Pearson/Prentice Hall) ISBN 0131139568; 9780131139565

\bibitem{FH_ruleofthumb}
Middlestead R~W 2017 {\em Digital Communications with Emphasis on Data Modems: Theory, Analysis, Design, Simulation, Testing, and Applications\/} (John Wiley \& Sons)

\bibitem{Mesulam}
Steriade M, Gloor P, Llinás R, {Lopes da Silva} F and Mesulam M~M 1990 Basic mechanisms of cerebral rhythmic activities {\em Electroencephalography and Clinical Neurophysiology\/} \href{https://doi.org/https://doi.org/10.1016/0013-4694(90)90001-Z}{{\bf 76} 481--508} ISSN 0013-4694

\bibitem{rangaswamy}
Rangaswamy M, Porjesz B, Chorlian D~B, Wang K, Jones K~A, Bauer L~O, Rohrbaugh J, O’Connor S~J, Kuperman S, Reich T and Begleiter H 2002 Beta power in the eeg of alcoholics {\em Biological Psychiatry\/} \href{https://doi.org/https://doi.org/10.1016/S0006-3223(02)01362-8}{{\bf 52} 831--842} ISSN 0006-3223

\bibitem{globavg}
Lin M 2013 Network in network {\em arXiv preprint arXiv:1312.4400\/}

\bibitem{globavgtimeinv}
Zhao L and Zhang Z 2024 A improved pooling method for convolutional neural networks {\em Scientific Reports\/} \href{https://doi.org/10.1038/s41598-024-51258-6}{{\bf 14} 1589} ISSN 2045-2322

\bibitem{overfitpaper}
Bejani M~M and Ghatee M 2021 A systematic review on overfitting control in shallow and deep neural networks {\em Artificial Intelligence Review\/} \href{https://doi.org/10.1007/s10462-021-09975-1}{{\bf 54} 6391--6438} ISSN 1573-7462

\bibitem{MLRM}
McCullagh P 2019 {\em Generalized linear models\/} (Routledge)

\bibitem{multicoll}
Young D~S 2018 {\em Handbook of regression methods\/} (Chapman and Hall/CRC)

\bibitem{holm}
Holm S 1979 A simple sequentially rejective multiple test procedure {\em Scandinavian Journal of Statistics\/} {\bf 6} 65--70 ISSN 03036898, 14679469

\bibitem{FBFS1}
Kutner M~H, Nachtsheim C~J, Neter J and Li W 2005 {\em Applied linear statistical models\/} (McGraw-hill)

\bibitem{FBFS2}
Weisberg S 2005 Applied linear regression

\bibitem{VIF}
James G, Witten D, Hastie T, Tibshirani R {\em et~al.\/} 2013 {\em An introduction to statistical learning\/} vol 112 (Springer)

\bibitem{QCV}
Kokoska S and Zwillinger D 2000 {\em CRC standard probability and statistics tables and formulae\/} (Crc Press)

\bibitem{wong}
Wong A 2019 Netscore: Towards universal metrics for large-scale performance analysis of deep neural networks for practical on-device edge usage {\em Image Analysis and Recognition: 16th International Conference, ICIAR 2019, Waterloo, ON, Canada, August 27–29, 2019, Proceedings, Part II\/} (Berlin, Heidelberg: Springer-Verlag) p 15–26 ISBN 978-3-030-27271-5

\bibitem{Rossini}
Rossini P~M, Rossi S, Babiloni C and Polich J 2007 Clinical neurophysiology of aging brain: From normal aging to neurodegeneration {\em Progress in Neurobiology\/} \href{https://doi.org/10.1016/j.pneurobio.2007.07.010}{{\bf 83} 375--400}

\bibitem{eegconformer}
Song Y, Zheng Q, Liu B and Gao X 2023 Eeg conformer: Convolutional transformer for eeg decoding and visualization {\em IEEE Transactions on Neural Systems and Rehabilitation Engineering\/} \href{https://doi.org/10.1109/TNSRE.2022.3230250}{{\bf 31} 710--719}

\bibitem{distillation}
Hinton G, Vinyals O and Dean J 2015 Distilling the knowledge in a neural network (\textit{Preprint} \eprint{1503.02531}) \url{https://arxiv.org/abs/1503.02531}

\bibitem{insuffEEG}
Dauwels J, Vialatte F and Cichocki A 2010 Diagnosis of alzheimers disease from eeg signals: Where are we standing? {\em Current Alzheimer Research\/} \href{https://doi.org/10.2174/156720510792231720}{{\bf 7} 487--505} ISSN 1567-2050/1875-5828

\bibitem{babiloni}
Babiloni C, Lizio R, Marzano N, Capotosto P, Soricelli A, Triggiani A~I, Cordone S, Gesualdo L and {Del Percio} C 2016 Brain neural synchronization and functional coupling in alzheimer's disease as revealed by resting state eeg rhythms {\em International Journal of Psychophysiology\/} \href{https://doi.org/https://doi.org/10.1016/j.ijpsycho.2015.02.008}{{\bf 103} 88--102} ISSN 0167-8760 research on Brain Oscillations and Connectivity in A New Take-Off State

\bibitem{UPDRS}
Goetz C~G, Fahn S, Martinez-Martin P, Poewe W, Sampaio C, Stebbins G~T, Stern M~B, Tilley B~C, Dodel R, Dubois B, Holloway R, Jankovic J, Kulisevsky J, Lang A~E, Lees A, Leurgans S, LeWitt P~A, Nyenhuis D, Olanow C~W, Rascol O, Schrag A, Teresi J~A, Van~Hilten J~J and LaPelle N 2007 Movement disorder society-sponsored revision of the unified parkinson's disease rating scale (mds-updrs): Process, format, and clinimetric testing plan {\em Movement Disorders\/} \href{https://doi.org/https://doi.org/10.1002/mds.21198}{{\bf 22} 41--47}

\bibitem{MMSE}
Folstein M~F, Folstein S~E and McHugh P~R 1975 “mini-mental state”: A practical method for grading the cognitive state of patients for the clinician {\em Journal of Psychiatric Research\/} \href{https://doi.org/https://doi.org/10.1016/0022-3956(75)90026-6}{{\bf 12} 189--198} ISSN 0022-3956

\bibitem{MOCA}
Nasreddine Z~S, Phillips N~A, Bédirian V, Charbonneau S, Whitehead V, Collin I, Cummings J~L and Chertkow H 2005 The montreal cognitive assessment, moca: A brief screening tool for mild cognitive impairment {\em Journal of the American Geriatrics Society\/} \href{https://doi.org/https://doi.org/10.1111/j.1532-5415.2005.53221.x}{{\bf 53} 695--699}

\bibitem{multimodalEEG}
Polikar R, Tilley C, Hillis B and Clark C~M 2010 Multimodal eeg, mri and pet data fusion for alzheimer's disease diagnosis {\em 2010 Annual International Conference of the IEEE Engineering in Medicine and Biology\/} pp 6058--6061

\bibitem{multimodalEEG1}
Cassani R, Estarellas M, San-Martin R, Fraga F~J and Falk T~H 2018 Systematic review on {Resting-State} {EEG} for alzheimer's disease diagnosis and progression assessment {\em Dis Markers\/} \href{https://doi.org/doi.org/10.1155/2018/5174815}{{\bf 2018} 5174815}

\end{thebibliography}

\end{document}